\documentclass[10pt,twocolumn,letterpaper]{article}

\usepackage{soul}
\usepackage{amsmath,amssymb}
\usepackage{comment}

\usepackage{algorithm}
\usepackage{algpseudocode}
\usepackage{booktabs}
\usepackage[shortlabels,inline]{enumitem}
\usepackage{multirow}
\usepackage{etoolbox}
\usepackage{float}
\usepackage[table]{xcolor}
\usepackage{xcolor}
\definecolor{Green}{rgb}{0.0,0.91,0.0}
\definecolor{Red}{rgb}{0.91,0.0,0.0}
\definecolor{Blue}{rgb}{0.0,0.0,0.91}
\usepackage[accsupp]{axessibility}

\usepackage{wacv}

\usepackage{titletoc}
\usepackage[toc,page]{appendix}

\definecolor{wacvblue}{rgb}{0.21,0.49,0.74}
\usepackage[pagebackref,breaklinks,colorlinks,allcolors=wacvblue]{hyperref}

\def\wacvPaperID{1105} % *** Enter the WACV Paper ID here
\def\confName{WACV}
\def\confYear{2027}

\title{\includegraphics[height=2.5ex]{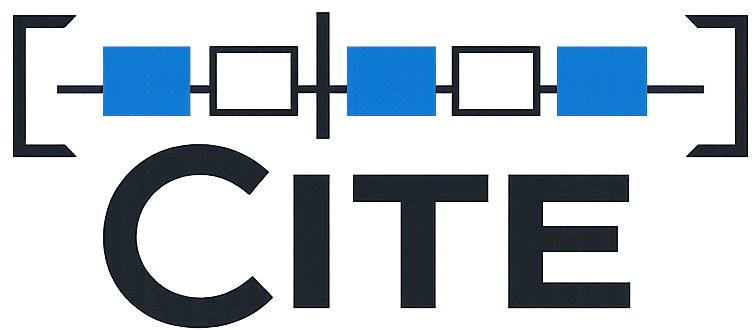}: Complementary and Irreplaceable Temporal Evidence for Keyframe Selection}

\author{
Awet Araia$^{\dag}$ \and
Mohamad Alansari$^{\dag}$ \and
Yonathan Michael$^{\dag}$ \and
Dwarikanath Mahapatra \and
Andreas Henschel \and
Naoufel Werghi \and
Sajid Javed\\[0.4em]
Department of Computer Science, Khalifa University, Abu Dhabi, UAE\\
{\tt\scriptsize [100043380,100061914,100053679,dwarikanath.mahapatra, andreas.henschel,naoufel.werghi,sajid.javed]@ku.ac.ae}\\
{\scriptsize $^{\dag}$Equal contribution}
}

\vspace{-5em}
\makeatletter
\apptocmd{\@maketitle}{%

    \begin{center}
        \vspace{-2em}\includegraphics[width=0.8\textwidth]{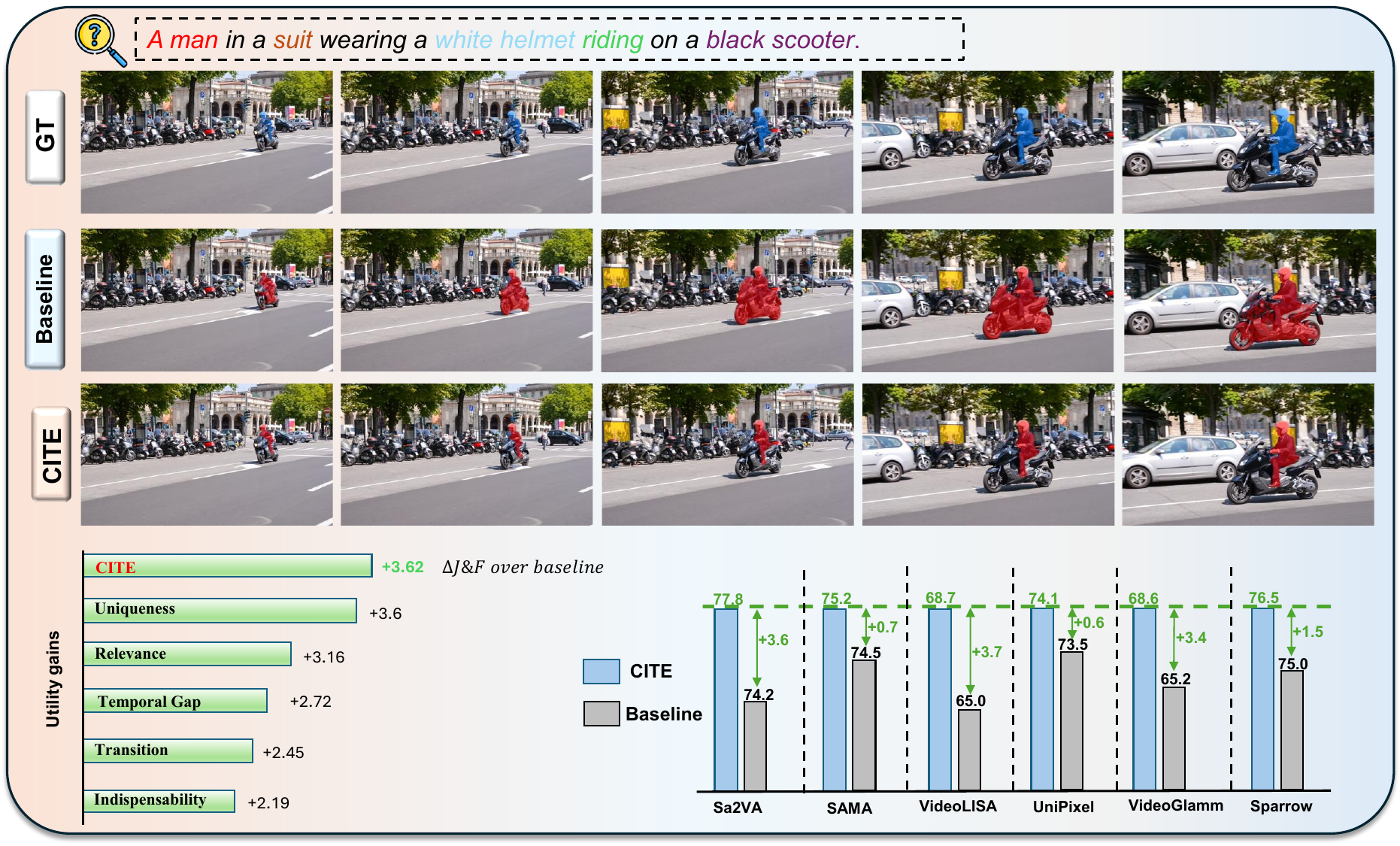}
        \begin{minipage}{\textwidth}
        \small
        \refstepcounter{figure}
        \label{fig:intro}
        Figure~\thefigure. \textbf{CITE improves RVOS by preserving query-specific evidence.} On Ref-DAVIS17, utility gains are evaluated using Sa2VA-4B, while backbone gains represent the average improvement in $\mathcal{J}{\&}\mathcal{F}$ over the corresponding baselines across all evaluated model sizes.
        \end{minipage}
    \end{center}
    \vspace{-0.5em}
}{}{}

\makeatother

\makeatletter
\patchcmd{\abstract}
  {\centerline{\large\bf Abstract}}
  {\centerline{\large\bf Abstract}\vspace{-0.6em}}
  {}{}
\makeatother

\begin{document}
\maketitle

\begin{abstract}
\noindent Pixel-grounded video-language models enable referring video object segmentation (RVOS), but dense-frame inference is costly, so most systems reason from sparse frames. Existing samplers emphasize uniform coverage or global query-frame relevance and may miss brief evidence for attributes, actions, motion, and spatial relations. We introduce CITE, a training-free, plug-and-play keyframe selector that frames sparse RVOS as query-conditioned complementary frames evidence preservation. CITE decomposes each expression into semantic and motion factors, estimates their framewise support using frozen vision-language features and lightweight temporal cues, and greedily selects five non-redundant frames that balance factor coverage, evidence uniqueness, and visual-temporal diversity. Without modifying or retraining the downstream model, CITE improves full-video grounding across multiple pixel-grounded RVOS backbones and benchmarks, with the largest gains for expressions containing sparse or short-lived discriminative cues. These results establish evidence-preserving selection as an effective route to efficient RVOS, while exposing limitations when dense temporal continuity is essential.
% Code is available at 
Code is at \url{https://github.com/HamadYA/CITE}.
\end{abstract}
    
% \vspace{-1.5em}
\vspace{-2em}
\section{Introduction}
\label{sec:intro}
\noindent Multi-modal large language models (MLLMs) have advanced language-conditioned video understanding~\cite{videollava,videoglamm,videolisa,videomolmo,sparrow,sa2va,unipixel,sama}. Pixel-grounded variants support referring video object segmentation (RVOS), localizing referred objects across time~\cite{videolisa,videoglamm,sparrow,sa2va,unipixel,sama,Instance-segmentation,interRVOS}. Because visual-token cost grows with the number of frames~\cite{moviechat}, practical systems reason from sparse subsets.

Existing Key-frame selection (KFS) methods prioritize temporal uniformity, saliency, visual diversity, global query-frame relevance, or combinations thereof~\cite{aks,qframe,adardkey,air,focus,frameoracle,videomme}. Although effective for coarse video understanding, these criteria can miss the localized, compositional evidence required by RVOS~\cite{videochatgpt,internvideo2,llavavideo,davis,mevis,ref_youtube_vos,revos,sa2va,MOSE}. A query may depend on category, appearance, action, motion, and spatial relations whose evidence occurs at different times~\cite{reason}. For example, identifying \emph{``the red car turning left behind the bus''} may require appearance evidence from one moment, motion evidence from another, and relational evidence from yet another. A frame may therefore be globally relevant yet omit the decisive disambiguating cue. Under a tight frame budget, selected frames must preserve complementary evidence rather than repeat what other frames already show. 

This motivates a complementary view of frame importance: \emph{if a frame were removed, would the remaining frames still suffice to resolve the query?} A frame is valuable when it provides evidence unavailable elsewhere. Directly estimating this value by removing each frame, rerunning RVOS inference, and measuring the drop in full-video segmentation quality is prohibitively expensive and backbone-specific. We therefore approximate frame irreplaceability before inference using lightweight, model-agnostic signals.
We propose \textbf{C}omplementary and \textbf{I}rreplaceable \textbf{T}emporal \textbf{E}vidence (\textbf{CITE}), a training-free, plug-and-play selector for evidence-preserving RVOS. We study RVOS under a minimal reasoning budget: unless otherwise stated, only $K=5$ selected frames are passed to the pixel-grounded video MLLM. Mask prediction and temporal propagation follow the model's original pipeline. Under this tight budget, missing an attribute, action, motion cue, or spatial relation can directly cause incorrect grounding because little redundant evidence remains. CITE decomposes each expression into semantic and motion cues, estimates their framewise support using frozen vision-language representations and lightweight temporal signals, and greedily selects five frames that balance cue coverage, evidence uniqueness, temporal sensitivity, and visual-temporal non-redundancy. The original expression and selected frames are passed to the unchanged downstream pixel-grounded video MLLM. Across multiple backbones and benchmarks, CITE improves full-video grounding, particularly when decisive evidence is sparse (see Fig.~\ref{fig:intro}).
\noindent Our contributions are:
\begin{itemize}[leftmargin=0pt, itemsep=0pt, topsep=2pt, parsep=0pt, partopsep=0pt]
    \item We formulate KFS under a minimal five-frame reasoning budget as \emph{query-conditioned evidence preservation}, with a complementary-frames objective balancing cue coverage, frame indispensability, and visual-temporal non-redundancy.
    \item We introduce CITE, a training-free, plug-and-play selector that decomposes referring expressions and estimates framewise semantic and motion evidence.
    \item We evaluate CITE across multiple pixel-grounded video MLLMs and RVOS benchmarks, analyzing when evidence-preserving selection improves downstream grounding.
\end{itemize}

\section{Related Work}
\label{sec:related_work}

\noindent \textbf{Pixel-grounded video MLLMs.}
Pixel-grounded video MLLMs extend holistic video understanding to language-guided object localization and segmentation~\cite{videolisa,sa2va,videoglamm,unipixel,sama,sparrow}. Despite differences in architectures, mask decoders, tracking, and instruction tuning, these models typically operate on constrained video inputs obtained through first-$N$, uniform, or fixed sparse--dense sampling~\cite{vidundsurv}. Such query-agnostic protocols may discard brief attributes, actions, motion changes, or spatial relations needed for fine-grained grounding. Orthogonal to model design, CITE selects compact, query-conditioned evidence frames before inference, reducing redundant inputs under the same frame budget.
\noindent \textbf{Keyframe selection.}
Because fixed samplers can miss brief, query-relevant events~\cite{kfsbench}, recent KFS methods adapt selection using relevance, temporal coverage, diversity, uncertainty, or computational budget. AKS~\cite{aks} balances query relevance and coverage under a token budget, whereas Q-Frame~\cite{qframe} performs training-free selection using CLIP-like image--text matching and multi-resolution adaptation. Other methods exploit large-scale text--video similarity~\cite{keyvideollm}, relevance--diversity trade-offs~\cite{adardkey}, adaptive budgets~\cite{frameoracle}, bandit exploration~\cite{focus}, or iterative reasoning~\cite{air}. Although these methods establish relevance and coverage as effective criteria, most assess the query globally or seek generic temporal representativeness. Fine-grained grounding instead requires complementary category, appearance, action, motion, and relational evidence that may occur at different times. CITE therefore decomposes the query, estimates framewise semantic and motion evidence, and selects a non-redundant subset balancing factor coverage, indispensability, and redundancy reduction.
\noindent \textbf{Trainable frame selection.}
Learnable selectors optimize frame retention rather than relying on frozen scoring signals. MLLM-Selector~\cite{MLLM-VfS} and FFS~\cite{ffs} use filtering modules or register tokens to estimate query-conditioned importance, whereas TSPO~\cite{tspo} trains an event-aware agent with GRPO and ReaSon~\cite{reason} applies a causal information bottleneck with counterfactual interventions. Agentic methods such as VideoAgent~\cite{videoagent} and VideoTree~\cite{videotree} instead iteratively ask an LLM which frames to inspect. Although effective for long-video reasoning, these approaches require selector-specific training that may couple the policy to a backbone or repeated LLM calls. CITE instead performs training-free, single-pass selection using lightweight semantic and motion cues before pixel-grounded inference, enabling transfer across frozen video MLLMs without selection-specific training or iterative calls.

% ===============================================================================================
\begin{figure*}[!t]
\centering
\includegraphics[width=0.65\textwidth]{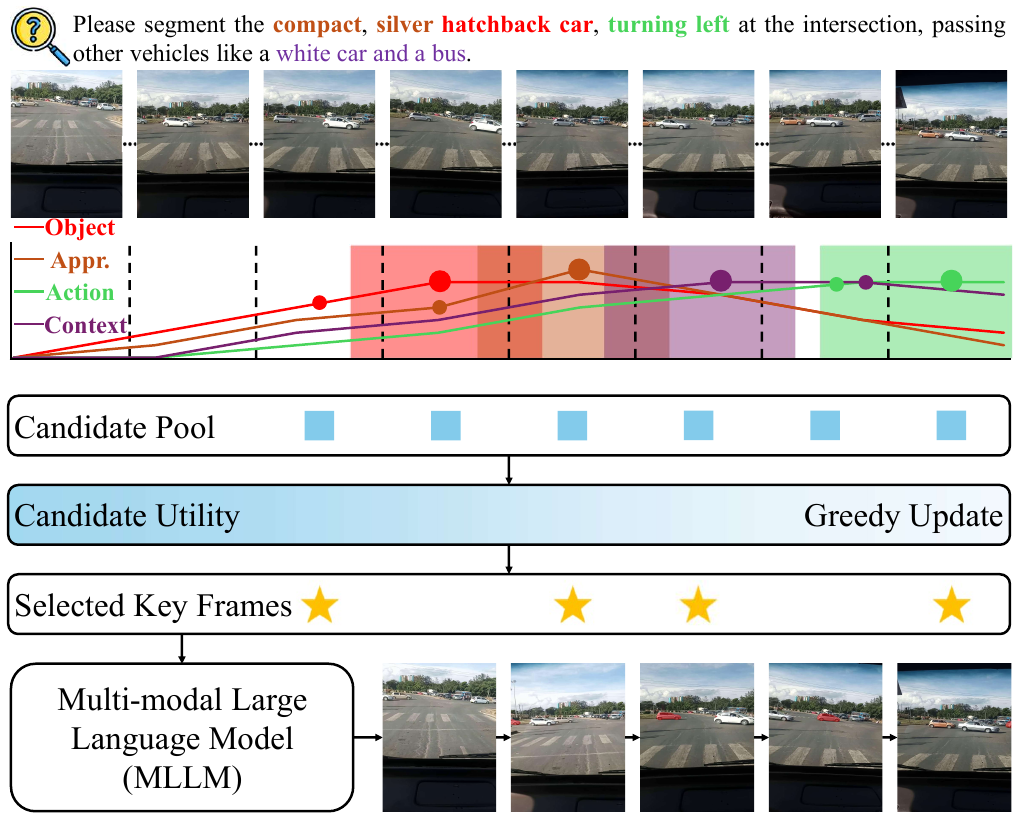}
\vspace{-1em}
\caption{
{\small
CITE factorizes the query into semantic and motion cues, then uses framewise evidence to select a
compact subset balancing utility, cue coverage, and visual--temporal redundancy.
}
}
\label{fig:main}
\vspace{-1em}
\end{figure*}
% ===============================================================================================

\section{CITE: {\normalsize Query-Conditioned Evidence-Preservation}}
\label{sec:method}

\noindent We propose CITE, a training-free KFS framework for RVOS. Given a video and a referring expression, CITE selects a compact set of frames before downstream inference, as illustrated in Fig.~\ref{fig:main} and Algo.~1. CITE is designed as a plug-and-play pre-processing module while keeping the original prompt format, mask decoder, tracking or propagation module, and all learned parameters of the grounding model unchanged.

\noindent \textbf{Problem formulation.}
Let $\mathcal{V}=\{x_t\}_{t=1}^{T}$ denote a video with $T$ frames, let $q$ be a referring expression, and let $K$ be the frame budget of the downstream model. The goal of KFS is to select a temporally ordered subset
% \begin{equation} 
%     S=\{s_1,\ldots,s_{K'}\}, \qquad
%     1\leq s_1<\cdots<s_{K'}\leq T,\qquad
%     K'\leq \min(K,T),
% \end{equation}
{\small
\begin{equation}
    S=\{s_1,\ldots,s_{K'}\}, \quad
    1\leq s_1<\cdots<s_{K'}\leq T,\quad
    K'\leq\min(K,T).
\end{equation}
}
that preserves the visual evidence needed to ground the referred object.
An ideal selector would choose frames whose absence would remove information necessary for the downstream RVOS grounding model. Let $f_{\theta}$ denote a frozen pixel-grounded video-language model and let $U_{\theta}(S;\mathcal{V},q)$ denote the usefulness of a selected subset $S$ for grounding $q$.
The ideal objective can be written as
\begin{equation} \small
    S^{*}
    =
    \arg\max_{S\subseteq \{1,\ldots,T\},\, |S|\leq K}
    U_{\theta}(S;\mathcal{V},q).
    \label{eq:ideal_objective}
\end{equation}
However, directly estimating Eq.~\eqref{eq:ideal_objective} would require repeatedly running the downstream RVOS model over many candidate subsets, which is computationally expensive and tightly coupled to a particular backbone. Instead, CITE optimizes a model-agnostic proxy before RVOS inference. The proxy favors frames that provide strong query evidence, support factors not already covered by selected frames, capture temporally localized changes, and are not easily replaceable by nearby visually similar frames.

\noindent \textbf{Query factorization.}
A referring expression often contains multiple grounding cues. 
For example, the expression ``the red car turning left behind the bus'' involves an object category, an appearance attribute, an action, a motion direction, and a spatial relation. 
A single global query-frame similarity score can collapse these cues and prefer frames that match the expression only partially. 
CITE therefore factorizes the expression into a small set of query factors:
\begin{equation} \small
    \mathcal{A}(q)=\{(a_m,g_m)\}_{m=1}^{M},
\end{equation}
where $a_m$ is the factor text and $g_m$ is its factor group. 
The groups include object nouns, colors, textures, materials, actions, motion directions, and coarse spatial relations. We use a lightweight deterministic extractor rather than a learned parser. The extractor identifies visual attributes and actions from lexical templates, detects directional and spatial phrases, and treats remaining informative tokens as candidate object nouns. 
This design keeps CITE training-free, deterministic, and independent of RVOS annotations. If no factor is extracted, CITE falls back to uniform sampling.

\noindent \textbf{Factor-level semantic evidence.}
For each non-directional factor $(a_m,g_m)$, CITE constructs a short text prompt $p_m$. 
Examples include ``a photo of \texttt{noun}'', ``a photo of a \texttt{color} object'', and ``a photo of something \texttt{action}''. 
Let $\phi_I$ and $\phi_T$ be the image and text encoders of a frozen vision-language model. 
We encode each frame and factor prompt as normalized embeddings:
\begin{equation} \small
    z_t = \frac{\phi_I(x_t)}{\|\phi_I(x_t)\|_2},
    \qquad
    h_m = \frac{\phi_T(p_m)}{\|\phi_T(p_m)\|_2}.
\end{equation}
The raw semantic evidence for factor $m$ at frame $t$ is the image-text similarity $\tilde e^{\rm sem}_m(t)=z_t^{\top}h_m.$ As score ranges vary across factors and videos, we normalize each evidence curve over time:
\begin{equation} \small
    e^{\rm sem}_m(t)
    =
    \operatorname{Norm}_{\tau\in\{1,\ldots,T\}}
    \left(\tilde e^{\rm sem}_m(\tau)\right),
    \label{eq:semantic_evidence}
\end{equation}
where $\operatorname{Norm}$ denotes min-max normalization within the same video and factor. 
This factor-level representation allows different frames to support different parts of the same query, rather than forcing all evidence to be captured by a single global query-frame score.

\noindent \textbf{Lightweight temporal evidence.}
Semantic similarity alone is often insufficient for expressions involving actions or motion directions. 
CITE therefore augments factor evidence with inexpensive temporal cues computed from frame differences. 
Let $G(x_t)$ denote the grayscale version of frame $x_t$. For $t>1$, we define
$D_t = |G(x_t)-G(x_{t-1})|, \quad D_1 = 0.$
The global motion magnitude is
\begin{equation} \small
    m(t)
    =
    \operatorname{Norm}_{\tau}
    \left(
    \sum_{i,j} D_t(i,j)
    \right).
    \label{eq:motion_magnitude}
\end{equation}
To obtain coarse direction evidence, we compute the center of frame difference:
\begin{equation} \small
    c_x(t)=
    \frac{\sum_{i,j} j\,D_t(i,j)}
    {\sum_{i,j}D_t(i,j)+\epsilon},
    \qquad
    c_y(t)=
    \frac{\sum_{i,j} i\,D_t(i,j)}
    {\sum_{i,j}D_t(i,j)+\epsilon}.
\end{equation}
The signed shifts
$\delta_x(t)=c_x(t)-c_x(t-1),\quad\delta_y(t)=c_y(t)-c_y(t-1)$
provide weak evidence for left/right and up/down motion. 
For a direction label $d$, CITE maps the corresponding signed shift to a nonnegative evidence curve. 
For example, rightward evidence uses $\max(\delta_x(t),0)$, while leftward evidence uses $\max(-\delta_x(t),0)$. 
We weight direction evidence by motion magnitude:
\begin{equation} \small
    e^{\rm dir}_d(t)
    =
    \operatorname{Norm}_{\tau}
    \left(
    \max(\sigma_d(t),0)\,(0.25+m(t))
    \right),
    \label{eq:direction_evidence}
\end{equation}
where $\sigma_d(t)$ is the signed shift associated with direction $d$.
For action factors, we combine semantic and motion evidence:
\begin{equation} \small
    e_m(t)
    =
    \operatorname{Norm}_{\tau}
    \left(
    0.7\,e^{\rm sem}_m(t) + 0.3\,m(t)
    \right).
    \label{eq:action_evidence}
\end{equation}
For non-action semantic factors, $e_m(t)=e^{\rm sem}_m(t)$; for direction
factors, $e_m(t)=e^{\rm dir}_m(t)$.
The motion cues $m(t)$ and $e^{\rm dir}_d(t)$ are not object-conditioned
optical flow or tracking signals; instead, they provide a weak,
model-agnostic signal for dynamic cues that may otherwise be missed under a
small frame budget.
The 0.7/0.3 weights in Eq.~\eqref{eq:action_evidence} are fixed across all
benchmarks and backbones; a sensitivity analysis of all coefficients is given
in Supp.\ Tabs. S15.
% 
% ===============================================================================================
\begin{algorithm}[h]
\caption{CITE keyframe selection (KFS)}
\label{alg:cite}
% \begin{minipage}{0.6\textwidth}
\begin{algorithmic}[1]
\Require Video frames $\mathcal{V}=\{x_t\}_{t=1}^{T}$, referring expression $q$, frame budget $K$, radius $r$
\State Extract query factors $\mathcal{A}(q)=\{(a_m,g_m)\}_{m=1}^{M}$
\State Encode all frames once with a frozen image-text encoder
\State Compute lightweight motion cues from frame differences
\If{$M=0$}
    \State \Return uniform samples
\EndIf
\For{each query factor $(a_m,g_m)$}
    \State Compute factor evidence curve $e_m(t)$ over all frames
\EndFor
\State Build candidate pool $\mathcal{C}$ using top-$L$ factor peaks and temporal expansion, Eq.~\eqref{eq:candidate_pool}
\State Compute standalone utility $u(t;q)$ for each $t\in\mathcal{C}$
\State $S\gets\emptyset$
\While{$|S|<\min(K,|\mathcal{C}|)$}
    \State Select $t^*$ by Eq.~\eqref{eq:greedy}
    \State $S\gets S\cup\{t^*\}$
\EndWhile
\State \Return temporally sorted $S$
\end{algorithmic}
% \end{minipage}
\end{algorithm}
% \vspace{-1em}
% ===============================================================================================
% 
\noindent \textbf{Evidence-focused candidate pool.}
Rather than greedily searching over all $T$ frames, CITE first builds a compact candidate pool from factor-level evidence peaks. 
For each factor $m$, we keep the top $L =\max(K,2)$
frames according to $e_m(t)$. 
We then expand each selected peak by a temporal radius $r$:
\begin{equation} \small
    \mathcal{C}
    =
    \bigcup_{m=1}^{M}
    \bigcup_{t\in \operatorname{Top}_{L}(e_m)}
    \{\tau: |\tau-t|\leq r,\; 1\leq \tau\leq T\}.
    \label{eq:candidate_pool}
\end{equation}
The neighborhood expansion allows frames adjacent to evidence peaks to compete during selection, which is important because frozen image-text scores can be temporally noisy. 
In our default setting, $r=1$.

\noindent \textbf{Complementary-frames evidence objective.}
CITE selects frames by maximizing a proxy for evidence preservation. 
The proxy has three parts: standalone frame utility, marginal factor coverage, and visual-temporal redundancy:
\begin{equation} \small
    \widehat U(S;q)
    =
    \sum_{t\in S} u(t;q)
    +
    \lambda_{\rm cov}
    \sum_{g\in\mathcal{G}} w_g
    \max_{t\in S}\max_{m:g_m=g} e_m(t)
    -
    \lambda_{\rm red}\Omega(S),
    \label{eq:set_objective}
\end{equation}
where $\mathcal{G}$ is the set of factor groups present in the query, $w_g$ is the group weight, $u(t;q)$ is the standalone utility of frame $t$, and $\Omega(S)$ penalizes redundant selected frames. 
The second term encourages the selected set to cover different types of query evidence, rather than repeatedly selecting frames that support the same factor.

\noindent \textit{Standalone frame utility.}
The utility $u(t;q)$ combines five evidence-loss proxies:
% {\footnotesize
% \begin{equation}  
%     u(t;q)=
%     \alpha_{\rm rel}u_{\rm rel}(t)
%     +
%     \alpha_{\rm uniq}u_{\rm uniq}(t)
%     +
%     \alpha_{\rm gap}u_{\rm gap}(t)
%     +
%     \alpha_{\rm trans}u_{\rm trans}(t)
%     +
%     \alpha_{\rm ind}u_{\rm ind}(t).
%     \label{eq:utility}
% \end{equation}
% }
\vspace{-1em}
\begin{equation} \small
\begin{aligned}
u(t;q)={}&
    \alpha_{\rm rel}u_{\rm rel}(t)
    +\alpha_{\rm uniq}u_{\rm uniq}(t)
    +\alpha_{\rm gap}u_{\rm gap}(t) \\
    &+\alpha_{\rm trans}u_{\rm trans}(t)
    +\alpha_{\rm ind}u_{\rm ind}(t).
\end{aligned}
\label{eq:utility}
\end{equation}

The relevance term measures average factor evidence:
$u_{\rm rel}(t)=\frac{1}{M}\sum_{m=1}^{M} e_m(t),\label{eq:rel_term}$
with a small additional motion contribution for expressions containing action or direction factors.
The uniqueness term estimates whether the evidence in frame $t$ can be replaced by a nearby candidate frame:
\begin{equation} \small
    u_{\rm uniq}(t)
    =
    \max
    \left(
    u_{\rm rel}(t)
    -
    \max_{\substack{t'\in\mathcal{C},\,t'\neq t\\ |t-t'|\leq w}}
    (z_t^{\top}z_{t'})u_{\rm rel}(t'),
    0
    \right),
    \label{eq:unique_term}
\end{equation}
where $w$ is a local temporal window. 
This term is high when a frame provides relevant evidence not well explained by visually similar neighboring candidates.
The gap term favors relevant frames located in sparse regions of the candidate pool using
$u_{\rm gap}(t)=\gamma(t)u_{\rm rel}(t),\label{eq:gap_term}$
where $\gamma(t)$ is the normalized temporal distance between the nearest candidate neighbors around $t$. 
This discourages the selector from over-concentrating on dense clusters of similar evidence peaks.
The transition term favors frames that capture motion or appearance changes as
$
    u_{\rm trans}(t)
    =
    0.4\,m(t)
    +
    0.4\,n(t)
    +
    0.2\,\Delta u_{\rm rel}(t),
    \label{eq:transition_term}
$
where $n(t)$ is local visual novelty, computed as one minus the average visual similarity between frame $t$ and its adjacent frames, and $\Delta u_{\rm rel}(t)$ is the local change in relevance score. 
This term helps preserve short-lived actions and appearance transitions.
The factor-indispensability term measures whether frame $t$ is among the strongest observations for individual factors using
$
    u_{\rm ind}(t)
    =
    \frac{1}{M}
    \sum_{m=1}^{M}
    \mathbb{I}
    \left[
    t\in \operatorname{Top}_{2}(e_m)
    \right].
    \label{eq:ind_term}
$
This term encourages frames that are especially important for rare or localized query factors.
In all experiments, we use fixed weights across datasets and backbones:
{\small
\begin{equation} \small
    (\alpha_{\rm rel},\alpha_{\rm uniq},\alpha_{\rm gap},
    \alpha_{\rm trans},\alpha_{\rm ind})
    =
    (0.20,0.20,0.30,0.15,0.15).
\end{equation}
} 

\noindent \textbf{Greedy evidence-preserving selection.}
We greedily maximize Eq.~\eqref{eq:set_objective} over the candidate pool. 
Starting from $S=\emptyset$, CITE repeatedly selects the candidate with the largest marginal score:
\begin{equation} \small
    t^{*}
    =
    \arg\max_{t\in\mathcal{C}\setminus S}
    \left[
    u(t;q)
    +
    \lambda_{\rm cov} C(t\mid S;q)
    -
    \lambda_{\rm red} R(t,S)
    \right].
    \label{eq:greedy}
\end{equation}
The coverage gain $C(t\mid S;q)$ measures how much adding $t$ improves the best evidence already covered for each factor group. 
For group $g$, define
\begin{equation} \small
    B_g(S)
    =
    \max_{s\in S}\max_{m:g_m=g} e_m(s),
    \qquad
    B_g(\emptyset)=0.
\end{equation}
Then
\begin{equation} \small
    C(t\mid S;q)
    =
    \sum_{g\in\mathcal{G}}
    w_g
    \max
    \left(
    B_g(S\cup\{t\})-B_g(S),
    0
    \right).
    \label{eq:coverage_gain}
\end{equation}
We use equal group weights, $w_g=1/|\mathcal{G}|$.
The redundancy penalty discourages selecting frames that are too close in time or too similar in appearance to frames already selected:
\begin{equation} \small
    R(t,S)
    =
    0.6
    \frac{1}{1+\min_{s\in S}|t-s|}
    +
    0.4
    \max_{s\in S} z_t^{\top}z_s,
    \label{eq:redundancy}
\end{equation}
with $R(t,\emptyset)=0$. We use $\lambda_{\rm cov}=0.25$ and $\lambda_{\rm red}=0.40$. 
After greedy selection, the chosen indices are sorted in temporal order before being passed to the downstream RVOS model.

\noindent \textbf{Plug-and-play inference and complexity.}
CITE runs as a pre-inference selector. For each video, frozen image embeddings and frame-difference cues are cached once and reused across all associated expressions. For each expression, CITE only encodes the extracted factor prompts, computes factor-frame evidence, builds $\mathcal{C}$, and performs greedy selection. The downstream RVOS model is otherwise unchanged as it receives the original expression and the selected frames in temporal order for reasoning. Thus, the dominant overhead is one frozen image-encoder pass per video, while the per-query cost is limited to $M$ text encodings, a small similarity matrix multiplication, and greedy search over $\mathcal{C}$. This makes CITE plug-and-play across frozen pixel-grounded video-language backbones.

% CITE is applied before downstream RVOS inference. 
% For each video, image embeddings and motion cues are computed once and reused for all referring expressions associated with that video. 
% For each expression, CITE only computes text embeddings for the extracted factors, forms evidence curves, constructs the candidate pool, and performs greedy selection. 
% The downstream model receives the original natural-language expression and the selected frames at their original timestamps for reasoning.
% % 
% The dominant additional cost is a single frozen image-encoder pass over the video. 
% The per-query cost is lightweight: $M$ text encodings, matrix multiplication between factor text embeddings and cached frame embeddings, simple frame-difference operations, and greedy search over the compact candidate pool. 
% Since the downstream RVOS model is unchanged, CITE can be used with different pixel-grounded video-language backbones without retraining or modifying their inference pipelines.

\noindent \textit{Approximation limits.}
CITE is an evidence-preservation proxy, not a direct estimate of each frame's causal effect on the final masks. It relies on frozen image-text scores and global frame differences, so distractors, camera motion, background changes, or parser errors can affect selection. We analyze these limitations in Sec.~\ref{sec:experiments}.

% % =======================================================================
% \begin{table}[!t]
% \centering
% \caption{
% {\small
% Runtime analysis. Timing reports selector overhead only, measured on a single A100 GPU with batch size 32 and averaged over 100-frame videos with 5 expressions per video.
% }
% }
% \vspace{-1em}
% \label{tab:runtime_overhead}
% \resizebox{\linewidth}{!}{
% \begin{tabular}{l c c c c c}
% \toprule
% Method 
% & Selector image pass
% & Per-video cost
% & Per-query cost
% & Frames to RVOS
% & Frame ratio to 5f \\
% \midrule
% Uniform
% & None 
% & 0.00s 
% & $<0.01$s
% & 5 
% & \(1.0\times\) \\

% Global-CLIP 
% & CLIP 
% & 1.05s
% & 0.01s 
% & 5 
% & \(1.0\times\) \\

% Q-Frame 
% & CLIP 
% & 1.18s
% & 0.02s 
% & 5 
% & \(1.0\times\) \\

% CITE
% & CLIP 
% & 1.21s 
% & 0.03s
% & 5 
% & \(1.0\times\) \\
% \midrule
% VideoLISA 
% & None 
% & 0.00s
% & 0.00s 
% & 36 
% & \(7.2\times\) \\

% VideoGLaMM 
% & None 
% & 0.00s 
% & 0.00s
% & 16 
% & \(3.2\times\) \\

% UniPixel/SPARROW 
% & None 
% & 0.00s
% & 0.00s
% & 16 
% & \(3.2\times\) \\

% Sa2VA/SAMA 
% & None 
% & 0.00s
% & 0.00s 
% & 5 
% & \(1.0\times\) \\
% \bottomrule
% \end{tabular}
% }
% \vspace{-1em}
% \end{table}
% % =======================================================================
% 
% 
% =======================================================================
\begin{table*}[!t]
\centering
\caption{
{\small
RVOS performance across pixel-grounded video MLLMs. Baselines use each model's official frame budget, while CITE uses 5 selected frames for all backbones. Deltas are computed against the corresponding official baseline.
}
}
\vspace{-1em}
\label{tab:main_results}
\resizebox{\textwidth}{!}{
\begin{tabular}{l c ccc ccc ccc ccc ccc}
\toprule
\multirow{2}{*}{Backbone} & \multirow{2}{*}{Venue}
& \multicolumn{3}{c}{Ref-DAVIS17}
& \multicolumn{3}{c}{MeViS \texttt{val$^\texttt{u}$}}
& \multicolumn{3}{c}{MeViS \texttt{val}}
& \multicolumn{3}{c}{Ref-YouTube-VOS}
& \multicolumn{3}{c}{Avg.} \\

\cmidrule(lr){3-5}
\cmidrule(lr){6-8}
\cmidrule(lr){9-11}
\cmidrule(lr){12-14}
\cmidrule(lr){15-17}

& 
& $\mathcal{J}$ & $\mathcal{F}$ & $\mathcal{J\&F}$
& $\mathcal{J}$ & $\mathcal{F}$ & $\mathcal{J\&F}$
& $\mathcal{J}$ & $\mathcal{F}$ & $\mathcal{J\&F}$
& $\mathcal{J}$ & $\mathcal{F}$ & $\mathcal{J\&F}$
& $\mathcal{J}$ & $\mathcal{F}$ & $\mathcal{J\&F}$ \\
\midrule

Sa2VA-1B \cite{sa2va} &
& 63.7 & 73.1 & 68.4 % DAVIS
& 46.5  & 54.9 & 50.7 % MeViS-U
& 40.1 & 46.3 & 43.2 % MeViS
& 66.1 & 69.7 & 67.9 % Ref-YT
& 54.1 & 61.0 & 57.6 % Avg
\\ 

\rowcolor{blue!7.5} + CITE & \multirow{-2}{*}{{\small\textcolor[gray]{0.15}{arXiv'25}}}
& 69.7 & 78.4 & 74.0\scriptsize{(\textcolor{red}{\textbf{5.6$\uparrow$}})}
& 48.7 & 57.1 & 52.9\scriptsize{(\textcolor{red}{\textbf{2.2$\uparrow$}})}
& 40.6 & 46.8 & 43.7\scriptsize{(\textcolor{red}{\textbf{0.5$\uparrow$}})}
& 66.4 & 70.1 & 68.2\scriptsize{(\textcolor{red}{\textbf{0.3$\uparrow$}})}
& 56.4 & 63.1 & 59.7\scriptsize{(\textcolor{red}{\textbf{2.1$\uparrow$}})}
\\

\midrule

Sa2VA-4B \cite{sa2va}&
& 69.4 & 77.7 & 73.6
& 48.6 & 56.8 & 52.7
& 38.9 & 44.7 & 41.8
& 69.5 & 73.4 & 71.5
& 56.6 & 63.2 & 59.9 % Avg
\\ 

\rowcolor{blue!7.5} + CITE & \multirow{-2}{*}{{\small\textcolor[gray]{0.15}{arXiv'25}}}
& 72.9 & 81.5 & 77.2\scriptsize{(\textcolor{red}{\textbf{3.6$\uparrow$}})}
& 51.8 & 59.1 & 55.4\scriptsize{(\textcolor{red}{\textbf{2.7$\uparrow$}})}
& 42.4 & 48.5 & 45.4\scriptsize{(\textcolor{red}{\textbf{3.6$\uparrow$}})}
& 70.0 & 74.0 & 72.0\scriptsize{(\textcolor{red}{\textbf{0.5$\uparrow$}})}
& 59.3 & 65.8 & 62.5\scriptsize{(\textcolor{red}{\textbf{2.6$\uparrow$}})}
\\ 

\midrule

Sa2VA-8B \cite{sa2va}&
& 72.1 & 81.0 & 76.6
& 52.7 & 61.0 & 56.8
& 44.3 & 50.8 & 47.6
& 70.1 & 74.6 & 72.4
& 59.8 & 66.9 & 63.3 % Avg
\\ 

\rowcolor{blue!7.5} + CITE & \multirow{-2}{*}{{\small\textcolor[gray]{0.15}{arXiv'25}}}
& 73.9 & 83.6 & 78.7\scriptsize{(\textcolor{red}{\textbf{2.1$\uparrow$}})}
& 55.9 & 63.8 & 59.8\scriptsize{(\textcolor{red}{\textbf{3.0$\uparrow$}})}
& 47.5 & 53.8 & 50.7\scriptsize{(\textcolor{red}{\textbf{3.1$\uparrow$}})}
& 70.4 & 74.8 & 72.6\scriptsize{(\textcolor{red}{\textbf{0.2$\uparrow$}})}
& 61.9 & 69.0 & 65.5\scriptsize{(\textcolor{red}{\textbf{2.2$\uparrow$}})} % Avg
\\ 

\midrule

Sa2VA-26B \cite{sa2va}&
& 73.4 & 82.5 & 78.0
& 53.9 & 63.3 & 58.6
& 43.0 & 50.1 & 46.6
& 72.9 & 77.5 & 75.2
& 60.8 & 68.4 & 64.6 % Avg
\\ 

\rowcolor{blue!7.5} + CITE & \multirow{-2}{*}{{\small\textcolor[gray]{0.15}{arXiv'25}}}
& 76.2 & 86.1 & 81.2\scriptsize{(\textcolor{red}{\textbf{3.2$\uparrow$}})}
& 55.6 & 64.3 & 60.0\scriptsize{(\textcolor{red}{\textbf{1.4$\uparrow$}})}
& 48.8 & 56.2 & 52.5\scriptsize{(\textcolor{red}{\textbf{5.9$\uparrow$}})}
& 75.4 & 79.8 & 77.6\scriptsize{(\textcolor{red}{\textbf{2.4$\uparrow$}})} % 74.68
& 64.0 & 71.6 & 67.8\scriptsize{(\textcolor{red}{\textbf{3.2$\uparrow$}})} % Avg
\\ 

\midrule

VideoLISA \cite{videolisa} &
& 60.9 & 69.2 & 65.0
& 46.9 & 53.3 & 50.1
& 36.6 & 42.2 & 39.4
& 59.9 & 62.5 & 61.2
& 51.1 & 56.8 & 53.9 % Avg
\\ 

\rowcolor{blue!7.5} + CITE & \multirow{-2}{*}{{\small\textcolor[gray]{0.15}{NeurIPS'24}}}
& 65.2 & 72.2 & 68.7\scriptsize{(\textcolor{red}{\textbf{3.7$\uparrow$}})}
& 45.2 & 52.0 & 48.6\scriptsize{(\textcolor{blue}{\textbf{1.5$\downarrow$}})}
& 36.7 & 42.3 & 39.5\scriptsize{(\textcolor{red}{\textbf{0.1$\uparrow$}})}
& 58.5 & 61.2 & 59.9\scriptsize{(\textcolor{blue}{\textbf{1.3$\downarrow$}})}
& 51.4 & 57.0 & 54.2\scriptsize{(\textcolor{red}{\textbf{0.3$\uparrow$}})} % Avg
\\

\midrule

SAMA-1B \cite{sama} &
& 68.5 & 77.3 & 72.9 % DAVIS
& 49.5 & 57.4 & 53.4 % MeViS-U
& 40.2 & 46.1 & 43.2 % MeViS
& 67.7 & 71.5 & 69.6 % Ref-YT
& 56.5 & 63.1 & 59.8 % Avg
\\ 

\rowcolor{blue!7.5} + CITE & \multirow{-2}{*}{{\small\textcolor[gray]{0.15}{NeurIPS'25}}}
& 68.9 & 77.8 & 73.3\scriptsize{(\textcolor{red}{\textbf{0.4$\uparrow$}})} % DAVIS
& 50.4 & 58.1 & 54.2\scriptsize{(\textcolor{red}{\textbf{0.8$\uparrow$}})} % MeViS-U
& 42.7 & 48.7 & 45.7\scriptsize{(\textcolor{red}{\textbf{2.5$\uparrow$}})} % MeViS
& 68.2 & 71.9 & 70.0\scriptsize{(\textcolor{red}{\textbf{0.4$\uparrow$}})} % Ref-YT
& 57.6 & 64.1 & 60.8\scriptsize{(\textcolor{red}{\textbf{1.0$\uparrow$}})} % Avg
\\

\midrule

SAMA-4B \cite{sama}&
& 69.8 & 78.2 & 74.0 % DAVIS
& 48.8 & 56.7 & 52.7 % MeViS-U
& 38.7 & 44.2 & 41.5 % MeViS
& 69.8 & 73.7 & 71.8 % Ref-YT
& 56.8 & 63.2 & 60.0 % Avg
\\ 

\rowcolor{blue!7.5} + CITE & \multirow{-2}{*}{{\small\textcolor[gray]{0.15}{NeurIPS'25}}}
& 72.1 & 80.6 & 76.4\scriptsize{(\textcolor{red}{\textbf{2.4$\uparrow$}})} % DAVIS
& 52.3 & 59.6 & 56.0\scriptsize{(\textcolor{red}{\textbf{3.3$\uparrow$}})} % MeViS-U
& 41.1 & 47.0 & 44.1\scriptsize{(\textcolor{red}{\textbf{2.6$\uparrow$}})} % MeViS
& 69.8 & 73.9 & 71.9\scriptsize{(\textcolor{red}{\textbf{0.1$\uparrow$}})} % Ref-YT
& 58.8 & 65.3 & 62.1\scriptsize{(\textcolor{red}{\textbf{2.1$\uparrow$}})} % Avg
\\ 

\midrule

SAMA-8B \cite{sama}&
& 72.2 & 81.4 & 76.8 % DAVIS
& 52.9 & 61.2 & 57.0 % MeViS-U
& 44.5 & 50.3 & 47.4 % MeViS
& 70.7 & 74.9 & 72.8 % Ref-YT
& 60.1 & 67.0 & 63.5 % Avg
\\ 

\rowcolor{blue!7.5} + CITE & \multirow{-2}{*}{{\small\textcolor[gray]{0.15}{NeurIPS'25}}}
& 73.2 & 82.5 & 77.9\scriptsize{(\textcolor{red}{\textbf{1.1$\uparrow$}})} % DAVIS
& 56.9 & 64.3 & 60.6\scriptsize{(\textcolor{red}{\textbf{3.6$\uparrow$}})} % MeViS-U
& 47.0 & 52.8 & 49.9\scriptsize{(\textcolor{red}{\textbf{2.5$\uparrow$}})} % MeViS
& 70.9 & 74.8 & 72.9\scriptsize{(\textcolor{red}{\textbf{0.1$\uparrow$}})} % Ref-YT
& 62.0 & 68.6 & 65.3\scriptsize{(\textcolor{red}{\textbf{1.8$\uparrow$}})} % Avg
\\

\midrule

UniPixel-3B \cite{unipixel} &
& 70.8 & 77.9 & 74.3
& 53.3 & 61.1 & 57.2
& 40.0 & 44.8 & 42.4
& 67.6 & 71.0 & 69.3
& 57.9 & 63.7 & 60.8 % Avg
\\ 

\rowcolor{blue!7.5} + CITE & \multirow{-2}{*}{{\small\textcolor[gray]{0.15}{NeurIPS'25}}}
& 70.7 & 78.9 & 74.8\scriptsize{(\textcolor{red}{\textbf{0.5$\uparrow$}})}
& 56.7 & 63.9 & 60.3\scriptsize{(\textcolor{red}{\textbf{3.1$\uparrow$}})}
& 43.6 & 48.6 & 46.1\scriptsize{(\textcolor{red}{\textbf{3.7$\uparrow$}})}
% &  &  & \hl{68.78}
& 68.9 & 72.2 & 70.6\scriptsize{(\textcolor{red}{\textbf{1.3$\uparrow$}})}
& 60.0 & 65.9 & 63.0\scriptsize{(\textcolor{red}{\textbf{2.2$\uparrow$}})} % Avg
\\ 

\midrule

UniPixel-7B \cite{unipixel} &
& 68.8 & 76.8 & 72.8
& 55.6 & 62.8 & 59.2
& 43.8 & 48.9 & 46.3
& 67.0 & 69.6 & 68.3
& 58.8 & 64.5 & 61.7 % Avg
\\ 

\rowcolor{blue!7.5} + CITE & \multirow{-2}{*}{{\small\textcolor[gray]{0.15}{NeurIPS'25}}}
& 69.2 & 77.7 & 73.5\scriptsize{(\textcolor{red}{\textbf{0.7$\uparrow$}})}
% &  &  & \hl{58.91}
& 57.1 & 64.0 & 60.6\scriptsize{(\textcolor{red}{\textbf{1.4$\uparrow$}})} % MeViS-U
& 45.1 & 50.1 & 47.6\scriptsize{(\textcolor{red}{\textbf{1.3$\uparrow$}})}
& 67.6 & 70.1 & 68.9\scriptsize{(\textcolor{red}{\textbf{0.6$\uparrow$}})}
& 59.8 & 65.5 & 62.7\scriptsize{(\textcolor{red}{\textbf{1.0$\uparrow$}})} % Avg
\\ 

\midrule

VideoGLaMM \cite{videoglamm}& 
& 59.8 & 70.6 & 65.2 % DAVIS
& 43.4 & 52.1 & 47.7 % MeViS-U
& 35.7 & 41.4 & 38.5 % MeVis
& 57.3 & 60.4 & 58.9 % Ref-YT
& 49.1 & 56.1 & 52.6 % Avg
\\ 

\rowcolor{blue!7.5} + CITE & \multirow{-2}{*}{{\small\textcolor[gray]{0.15}{CVPR'25}}}
& 64.4 & 72.6 & 68.6\scriptsize{(\textcolor{red}{\textbf{3.4$\uparrow$}})} % DAVIS
& 45.1 & 54.0 & 49.6\scriptsize{(\textcolor{red}{\textbf{1.9$\uparrow$}})} % MeViS-U
& 38.2 & 44.4 & 41.3\scriptsize{(\textcolor{red}{\textbf{2.8$\uparrow$}})} % MeVis
& 58.0 & 61.0 & 59.5\scriptsize{(\textcolor{red}{\textbf{0.6$\uparrow$}})} % Ref-YT
& 51.4 & 58.0 & 54.8\scriptsize{(\textcolor{red}{\textbf{2.2$\uparrow$}})} % Avg
\\ 

\midrule

SPARROW \cite{sparrow} & 
& 71.0 & 79.1 & 75.0 % DAVIS
& 42.8 & 49.5 & 46.1 % MeViS-U
& 34.5 & 40.0 & 37.3 % MeViS
& 48.4 & 49.0 & 48.7 % Ref-YT
& 49.2 & 54.4 & 51.8 % Avg
\\

\rowcolor{blue!7.5} + CITE & \multirow{-2}{*}{{\small\textcolor[gray]{0.15}{CVPR'26}}}
& 72.4 & 80.6 & 76.5\scriptsize{(\textcolor{red}{\textbf{1.5$\uparrow$}})} % DAVIS
& 43.4 & 50.3 & 46.8\scriptsize{(\textcolor{red}{\textbf{0.7$\uparrow$}})} % MeViS-U
& 37.1 & 43.1 & 40.1\scriptsize{(\textcolor{red}{\textbf{2.8$\uparrow$}})} % MeViS
& 49.5 & 50.2 & 49.9\scriptsize{(\textcolor{red}{\textbf{1.2$\uparrow$}})} % Ref-YT
& 50.6 & 56.1 & 53.3\scriptsize{(\textcolor{red}{\textbf{1.5$\uparrow$}})} % Avg
\\

\bottomrule
\end{tabular}
}
\vspace{-1em}
\end{table*}
% =======================================================================

% \vspace{-1em}
\section{Experiments}
\label{sec:experiments}
% \vspace{-1em}
\noindent We evaluate CITE as a plug-and-play selector for frozen pixel-grounded video MLLMs. All \(+\)CITE variants use \(K=5\) selected reasoning frames, while baselines follow our reproduction of each backbone's default pipeline. Thus, comparisons against denser defaults test aggressive frame compression, whereas comparisons against five-frame defaults isolate the effect of replacing the sampler.

% \vspace{-1em}
\subsection{Experimental Setup}

\noindent \textbf{Benchmarks and metrics.}
We evaluate Ref-DAVIS17 \texttt{val}~\cite{davis}, MeViS \texttt{val} and \texttt{val$^\texttt{u}$}~\cite{mevis}, Ref-YouTube-VOS \texttt{val}~\cite{ref_youtube_vos}, ReVOS~\cite{revos}, and Ref-SAV~\cite{sa2va}, covering standard, motion-centric, reasoning-oriented, and short/long-video RVOS. Following official protocols, we report region similarity ($\mathcal{J}$), contour accuracy ($\mathcal{F}$), and their mean ($\mathcal{J}\&\mathcal{F}$); tables show ($\mathcal{J}\&\mathcal{F}$) unless stated otherwise.
\noindent \textbf{Backbones.}
We apply CITE to frozen VideoLISA~\cite{videolisa}, VideoGLaMM~\cite{videoglamm}, Sa2VA~\cite{sa2va}, UniPixel~\cite{unipixel}, SAMA~\cite{sama}, and SPARROW~\cite{sparrow}. Only frame selection changes; downstream prompts, mask decoders, tracking or propagation modules, and learned parameters remain unchanged.
\noindent \textbf{Frame protocol.}
Baseline reproductions retain each model's native sampler and budget: 36 frames for VideoLISA; 16 uniformly sampled frames for VideoGLaMM, SPARROW, and UniPixel; and the first five frames for Sa2VA and SAMA. All (+)CITE variants instead use (K=5) selected frames. Comparisons with denser defaults therefore test five-frame compression, whereas those with five-frame defaults isolate sampler replacement.
\noindent \textbf{KFS baselines.}
For a controlled same-budget comparison, we evaluate Uniform, Global-CLIP~\cite{clip}, AKS~\cite{aks}, Q-Frame~\cite{qframe}, and CITE with (K=5). Global-CLIP ranks frames by CLIP similarity to the full expression and temporally orders the top (K). Official default results appear only as references; all selector comparisons use five frames, isolating the selection rule from the input budget.
\noindent \textbf{Implementation.}
Following~\cite{aks,qframe}, CITE uses frozen CLIP~\cite{clip} ViT-L/14. Unless stated otherwise, we fix (K=5), candidate expansion radius (r=1), and the selection weights in Sec.~\ref{sec:method} across datasets, backbones, and model scales. Frame embeddings and motion cues are cached once per video and reused across its expressions. 
\noindent \textbf{Runtime accounting.}
On a single A100 GPU (batch size 32), CITE requires 1.21,s per 100-frame video and 0.03,s per expression, averaged over five expressions per video. Corresponding video/expression costs are 1.05/0.01,s for Global-CLIP and 1.18/0.02,s for Q-Frame; Uniform is negligible. Timings exclude downstream segmentation and propagation; the CLIP image pass is cached across expressions. All selectors pass five frames to the RVOS model.
% 
% =======================================================================
\begin{table*}[!t]
\centering
\caption{
{\small
Results on ReVOS and Ref-SAV. Baseline rows use each model's official frame budget, while all \(+\)CITE rows use five selected evidence frames. Deltas are computed against the corresponding official baseline.
}
}
\vspace{-1em}
\label{tab:revos}
\resizebox{\textwidth}{!}{
\begin{tabular}{l ccc ccc ccc ccc ccc ccc ccc ccc ccc}

\toprule

\multirow{3}{*}{Backbone}
& \multicolumn{9}{c}{ReVOS}
& \multicolumn{9}{c}{Ref-SAV}
\\

& \multicolumn{3}{c}{Referring}
& \multicolumn{3}{c}{Reasoning}
& \multicolumn{3}{c}{Overall}

& \multicolumn{3}{c}{Long}
& \multicolumn{3}{c}{Short}
& \multicolumn{3}{c}{Overall}
\\

\cmidrule(lr){2-4}
\cmidrule(lr){5-7}
\cmidrule(lr){8-10}
\cmidrule(lr){11-13}
\cmidrule(lr){14-16}
\cmidrule(lr){17-19}

& $\mathcal{J}$ & $\mathcal{F}$ & $\mathcal{J\&F}$
& $\mathcal{J}$ & $\mathcal{F}$ & $\mathcal{J\&F}$
& $\mathcal{J}$ & $\mathcal{F}$ & $\mathcal{J\&F}$
& $\mathcal{J}$ & $\mathcal{F}$ & $\mathcal{J\&F}$
& $\mathcal{J}$ & $\mathcal{F}$ & $\mathcal{J\&F}$
& $\mathcal{J}$ & $\mathcal{F}$ & $\mathcal{J\&F}$
\\ \midrule

Sa2VA-1B \cite{sa2va}
& 56.0 & 61.0 & 58.5 
& 44.9 & 51.1 & 48.0
& 50.5 & 56.1 & 53.3

& 65.0 & 67.8 & 66.4
& 48.6 & 51.5 & 50.0
& 56.8 & 59.6 & 58.2
\\ 

\rowcolor{blue!7.5} + CITE
& 55.9 & 61.2 & 58.6\scriptsize{(\textcolor{red}{\textbf{0.1$\uparrow$}})}
& 45.2 & 51.5 & 48.3\scriptsize{(\textcolor{red}{\textbf{0.3$\uparrow$}})}
& 50.6 & 56.3 & 53.5\scriptsize{(\textcolor{red}{\textbf{0.2$\uparrow$}})}

& 66.5 & 69.3 & 67.9\scriptsize{(\textcolor{red}{\textbf{1.5$\uparrow$}})}
& 49.8 & 52.0 & 50.9\scriptsize{(\textcolor{red}{\textbf{0.9$\uparrow$}})}
& 58.2 & 60.7 & 59.4\scriptsize{(\textcolor{red}{\textbf{1.2$\uparrow$}})}
\\

\midrule

Sa2VA-4B \cite{sa2va}
& 60.3 & 65.3 & 62.8 
& 53.3 & 59.3 & 56.3
& 56.8 & 62.3 & 59.5

& 71.9 & 74.6 & 73.3
& 56.8 & 59.4 & 58.1
& 64.4 & 67.0 & 65.7
\\

\rowcolor{blue!7.5} + CITE 
& 60.7 & 65.5 & 63.1\scriptsize{(\textcolor{red}{\textbf{0.3$\uparrow$}})}
& 54.1 & 60.1 & 57.1\scriptsize{(\textcolor{red}{\textbf{0.8$\uparrow$}})}
& 57.4 & 62.8 & 60.1\scriptsize{(\textcolor{red}{\textbf{0.6$\uparrow$}})}

& 73.4 & 75.9 & 74.7\scriptsize{(\textcolor{red}{\textbf{1.4$\uparrow$}})}
& 57.6 & 60.0 & 58.8\scriptsize{(\textcolor{red}{\textbf{0.7$\uparrow$}})}
& 65.5 & 67.9 & 66.7\scriptsize{(\textcolor{red}{\textbf{1.0$\uparrow$}})}
\\

\midrule

Sa2VA-8B \cite{sa2va}
& 62.4 & 68.0 & 65.2
& 55.8 & 62.4 & 59.1
& 59.1 & 65.2 & 62.1

& 71.2 & 74.2 & 72.7
& 55.7 & 58.8 & 57.2
& 63.5 & 66.5 & 65.0
\\ 

\rowcolor{blue!7.5} + CITE 
& 63.7 & 69.3 & 66.5\scriptsize{(\textcolor{red}{\textbf{1.3$\uparrow$}})}
& 56.9 & 63.7 & 60.3\scriptsize{(\textcolor{red}{\textbf{1.2$\uparrow$}})}
& 60.3 & 66.5 & 63.4\scriptsize{(\textcolor{red}{\textbf{1.3$\uparrow$}})}

& 73.0 & 75.7 & 74.3\scriptsize{(\textcolor{red}{\textbf{1.6$\uparrow$}})}
& 55.8 & 58.7 & 57.3\scriptsize{(\textcolor{red}{\textbf{0.1$\uparrow$}})}
& 64.4 & 67.1 & 65.7\scriptsize{(\textcolor{red}{\textbf{0.7$\uparrow$}})}
\\

\midrule

Sa2VA-26B \cite{sa2va}
& 64.4 & 69.9 & 67.1
& 56.3 & 63.1 & 59.7
& 60.3 & 66.5 & 63.4

& 72.9 & 76.0 & 74.4
& 58.1 & 61.0 & 59.6
& 65.5 & 68.5 & 67.0
\\ 

\rowcolor{blue!7.5} + CITE
& 65.9 & 71.5 & 68.7\scriptsize{(\textcolor{red}{\textbf{1.6$\uparrow$}})}
& 57.9 & 64.9 & 61.4\scriptsize{(\textcolor{red}{\textbf{1.7$\uparrow$}})}
& 61.9 & 68.2 & 65.0\scriptsize{(\textcolor{red}{\textbf{1.6$\uparrow$}})}

& 75.6 & 78.6 & 77.1\scriptsize{(\textcolor{red}{\textbf{2.7$\uparrow$}})}
& 61.0 & 63.7 & 62.4\scriptsize{(\textcolor{red}{\textbf{2.8$\uparrow$}})}
& 68.3 & 71.2 & 69.7\scriptsize{(\textcolor{red}{\textbf{2.7$\uparrow$}})}
\\

\midrule

VideoLISA \cite{videolisa}
& 50.5 & 54.5 & 52.5 
& 40.8 & 46.0 & 43.4
& 45.6 & 50.3 & 47.9

& 34.6 & 36.1 & 35.3
& 22.3 & 24.9 & 23.6
& 28.5 & 30.5 & 29.5
\\

\rowcolor{blue!7.5} + CITE 
& 49.7 & 53.8 & 51.7
& 40.4 & 45.6 & 43.0
& 45.0 & 49.7 & 47.3\scriptsize{(\textcolor{blue}{\textbf{0.6$\downarrow$}})}

& 33.7 & 35.2 & 34.5
& 21.3 & 23.5 & 22.4
& 27.5 & 29.4 & 28.4\scriptsize{(\textcolor{blue}{\textbf{1.1$\downarrow$}})}
\\ 

\midrule

SAMA-1B \cite{sama}
& 46.6 & 50.9 & 48.7
& 38.4 & 43.4 & 40.9 
& 42.5 & 47.1 & 44.8

& 70.7 & 73.4 & 72.1
& 56.3 & 59.0 & 57.6
& 63.5 & 66.2 & 64.9
\\ 

\rowcolor{blue!7.5} + CITE 
& 58.2 & 63.6 & 60.9\scriptsize{(\textcolor{red}{\textbf{12.2$\uparrow$}})}
& 48.5 & 54.7 & 51.6\scriptsize{(\textcolor{red}{\textbf{10.7$\uparrow$}})}
& 53.4 & 59.2 & 56.3\scriptsize{(\textcolor{red}{\textbf{11.5$\uparrow$}})}

& 72.9 & 75.6 & 74.3\scriptsize{(\textcolor{red}{\textbf{2.2$\uparrow$}})}
% & 55.4 & 58.1 & 56.7
& 57.8 & 60.5 & 59.2\scriptsize{(\textcolor{red}{\textbf{1.6$\uparrow$}})}
% & 65.1 & 66.8 & 65.5\scriptsize{(\textcolor{red}{\textbf{0.6$\uparrow$}})}
& 65.4 & 68.1 & 66.8\scriptsize{(\textcolor{red}{\textbf{1.9$\uparrow$}})}
\\

\midrule

SAMA-4B \cite{sama}
& 60.1 & 64.8 & 62.5
& 52.9 & 58.6 & 55.8
& 56.5 & 61.7 & 59.1

& 74.1 & 76.6 & 75.3 
& 58.8 & 61.2 & 60.0 
& 66.4 & 68.9 & 67.7
\\ 

\rowcolor{blue!7.5} + CITE
& 61.6 & 66.5 & 64.0\scriptsize{(\textcolor{red}{\textbf{1.5$\uparrow$}})}
& 55.1 & 61.2 & 58.1\scriptsize{(\textcolor{red}{\textbf{2.3$\uparrow$}})}
& 58.4 & 63.8 & 61.1\scriptsize{(\textcolor{red}{\textbf{2.0$\uparrow$}})}

& 75.5 & 78.2 & 76.9\scriptsize{(\textcolor{red}{\textbf{1.6$\uparrow$}})}
& 59.8 & 62.1 & 60.9\scriptsize{(\textcolor{red}{\textbf{0.9$\uparrow$}})}
& 67.7 & 70.1 & 68.9\scriptsize{(\textcolor{red}{\textbf{1.2$\uparrow$}})}
\\ 

\midrule

SAMA-8B \cite{sama}
& 62.8 & 67.8 & 65.3
& 55.8 & 61.7 & 58.8
& 59.3 & 64.8 & 62.0

& 74.2 & 77.0 & 75.6
& 57.5 & 60.3 & 58.9
& 65.8 & 68.6 & 67.2
\\ 

\rowcolor{blue!7.5} + CITE
& 63.9 & 69.1 & 66.5\scriptsize{(\textcolor{red}{\textbf{1.2$\uparrow$}})}
& 57.8 & 63.8 & 60.8\scriptsize{(\textcolor{red}{\textbf{2.0$\uparrow$}})}
& 60.9 & 66.4 & 63.6\scriptsize{(\textcolor{red}{\textbf{1.6$\uparrow$}})}

& 75.2 & 78.0 & 76.6\scriptsize{(\textcolor{red}{\textbf{1.0$\uparrow$}})}
& 57.9 & 60.8 & 59.4\scriptsize{(\textcolor{red}{\textbf{0.5$\uparrow$}})}
& 66.6 & 69.4 & 68.0\scriptsize{(\textcolor{red}{\textbf{0.8$\uparrow$}})}
\\

\midrule

UniPixel-3B \cite{unipixel}
& 62.7 & 67.1 & 64.9
& 55.7 & 60.7 & 58.2
& 59.2 & 63.9 & 61.5

& 66.3 & 68.4 & 67.3
& 54.4 & 57.2 & 55.8
& 60.4 & 62.8 & 61.6
\\ 

\rowcolor{blue!7.5} + CITE 
& 63.6 & 68.2 & 65.9\scriptsize{(\textcolor{red}{\textbf{1.0$\uparrow$}})}
& 57.4 & 62.8 & 60.1\scriptsize{(\textcolor{red}{\textbf{1.9$\uparrow$}})}
& 60.5 & 65.5 & 63.0\scriptsize{(\textcolor{red}{\textbf{1.5$\uparrow$}})}

& 67.9 & 70.5 & 69.2\scriptsize{(\textcolor{red}{\textbf{1.9$\uparrow$}})}
& 55.1 & 58.0 & 56.6\scriptsize{(\textcolor{red}{\textbf{0.8$\uparrow$}})}
& 61.5 & 64.3 & 62.9\scriptsize{(\textcolor{red}{\textbf{1.3$\uparrow$}})}
\\

\midrule

UniPixel-7B \cite{unipixel}
& 62.0 & 65.9 & 64.0
& 57.9 & 62.4 & 60.2
& 59.9 & 64.2 & 62.1

& 67.5 & 69.5 & 68.5
& 54.4 & 56.3 & 55.3
& 61.0 & 62.9 & 61.9
\\ 

\rowcolor{blue!7.5} + CITE
& 63.1 & 66.9 & 65.0\scriptsize{(\textcolor{red}{\textbf{1.0$\uparrow$}})}
& 59.6 & 64.1 & 61.9\scriptsize{(\textcolor{red}{\textbf{1.7$\uparrow$}})}
& 61.4 & 65.5 & 63.5\scriptsize{(\textcolor{red}{\textbf{1.4$\uparrow$}})}

& 69.4 & 71.6 & 70.5\scriptsize{(\textcolor{red}{\textbf{2.0$\uparrow$}})}
& 55.1 & 57.0 & 56.1\scriptsize{(\textcolor{red}{\textbf{0.8$\uparrow$}})}
& 62.3 & 64.3 & 63.3\scriptsize{(\textcolor{red}{\textbf{1.4$\uparrow$}})}
\\

\midrule

VideoGLaMM \cite{videoglamm}
& 30.6 & 33.6 & 32.1
& 21.8 & 24.9 & 23.4
& 26.2 & 29.3 & 27.8

& 42.2 & 43.9 & 43.1
& 45.8 & 48.0 & 46.9
& 44.0 & 46.0 & 45.0
\\ 

\rowcolor{blue!7.5} + CITE
& 47.1 & 52.5 & 49.8\scriptsize{(\textcolor{red}{\textbf{17.7$\uparrow$}})}
& 36.2 & 42.5 & 39.4\scriptsize{(\textcolor{red}{\textbf{16.0$\uparrow$}})}
& 41.7 & 47.5 & 44.6\scriptsize{(\textcolor{red}{\textbf{16.8$\uparrow$}})}

& 49.7 & 52.3 & 51.0\scriptsize{(\textcolor{red}{\textbf{7.9$\uparrow$}})}
& 47.2 & 49.4 & 48.3\scriptsize{(\textcolor{red}{\textbf{1.4$\uparrow$}})}
& 48.5 & 50.9 & 49.7\scriptsize{(\textcolor{red}{\textbf{4.7$\uparrow$}})}
\\

\midrule

SPARROW \cite{sparrow}
& 40.5 & 43.6 & 42.1
& 30.9 & 36.2 & 33.6
& 35.7 & 39.9 & 37.8

& 29.3 & 31.0 & 30.2
& 18.1 & 19.4 & 18.7
& 23.7 & 25.2 & 24.5
\\

\rowcolor{blue!7.5} + CITE
& 42.1 & 45.0 & 43.6\scriptsize{(\textcolor{red}{\textbf{1.5$\uparrow$}})}
& 32.7 & 37.9 & 35.3\scriptsize{(\textcolor{red}{\textbf{1.7$\uparrow$}})}
& 37.4 & 41.7 & 39.5\scriptsize{(\textcolor{red}{\textbf{1.7$\uparrow$}})}

& 29.8 & 31.7 & 30.8\scriptsize{(\textcolor{red}{\textbf{0.6$\uparrow$}})}
& 19.2 & 20.7 & 19.9\scriptsize{(\textcolor{red}{\textbf{1.2$\uparrow$}})}
& 24.5 & 26.2 & 25.4\scriptsize{(\textcolor{red}{\textbf{0.9$\uparrow$}})}
\\

\bottomrule
\end{tabular}
}
\vspace{-1em}
\end{table*}
% =======================================================================

% \vspace{-1em}
\subsection{Standard RVOS Benchmarks}
\vspace{-0.5em}
\noindent Tab.~\ref{tab:main_results} shows that five CITE frames improve average ($\mathcal{J}\&\mathcal{F}$) for most backbones. Across Sa2VA scales, gains range from (+2.1) to (+3.2), with the largest dataset gain of (+5.9) for Sa2VA-26B on MeViS \texttt{val}; gains on Ref-YouTube-VOS are generally smaller, consistent with more temporally persistent evidence. Comparisons with denser native inputs are compression results: VideoGLaMM improves after reducing 16 frames to five, whereas VideoLISA's 36-to-five (7.2$\times$) compression improves Ref-DAVIS17 but degrades MeViS \texttt{val$^\texttt{u}$} and Ref-YouTube-VOS, showing that some backbones benefit from dense temporal context.

% \vspace{-1em}
\subsection{Reasoning and Long-Video Grounding}
\vspace{-0.5em}
\noindent Tab.~\ref{tab:revos} shows that CITE improves ReVOS Overall ($\mathcal{J}\&\mathcal{F}$) for every backbone except VideoLISA, with the largest gains for VideoGLaMM (+16.8) and SAMA-1B (+11.5), supporting factor-level selection for compositional or temporally localized cues. On Ref-SAV, gains are smaller but positive for most backbones, led by VideoGLaMM (+4.7) and Sa2VA-26B (+2.7); VideoLISA decreases by (1.1), exposing the limitation of 5-frame compression when dense temporal continuity matters.

\vspace{-0.4em}
\subsection{Comparison with KFS Strategies}
\label{sec:kfs_comparison}
\vspace{-0.5em}
\noindent Under the controlled (K=5) comparison in Tab.~\ref{tab:kfs_comparison}, CITE achieves the best score for every dataset and backbone. It exceeds the strongest competing selector, Q-Frame~\cite{qframe}, by (0.2)-(0.7) Avg. points across backbones, indicating that complementary evidence is more effective than temporal coverage or global sentence-frame relevance alone.
\vspace{-0.6em}
% =======================================================================
\begin{table}[!h]
\centering
\caption{\small Controlled KFS comparison. Selector rows use (K=5); native-budget baselines are shown only for reference.}
\vspace{-1em}
\label{tab:kfs_comparison}
\resizebox{\linewidth}{!}{
\begin{tabular}{l l c c c c c}
\toprule
Backbone & Sampler 
& Ref-DAVIS17 
& MeViS \texttt{val$^\texttt{u}$}
& MeViS \texttt{val}
& Ref-YouTube-VOS
& Avg. \\
\midrule

\multirow{6}{*}{Sa2VA-1B}
& Baseline  & 68.4 & 50.7 & 43.2 & 67.9 & 57.6 \\
& Uniform  & 72.2 & 52.1 & 42.9 & 67.5 & 58.7 \\
& Global-CLIP  & 71.0 & 52.4 & 43.0 & 67.2 & 58.4 \\
& AKS  & 71.8 & 52.5 & 43.1 & 67.7 & 58.8 \\
& Q-Frame  & 72.6 & 52.7 & 43.4 & 68.0 & 59.2 \\
\rowcolor{blue!7.5}
& CITE  & \textbf{74.0} & \textbf{52.9} & \textbf{43.7} & \textbf{68.2} & \textbf{59.7} \\
\midrule

\multirow{6}{*}{Sa2VA-4B}
& Baseline  & 73.6 & 52.7 & 41.8 & 71.5 & 59.9 \\
& Uniform  & 76.4 & 54.8 & 44.8 & 71.6 & 61.9 \\
& Global-CLIP  & 75.9 & 55.0 & 44.4 & 71.3 & 61.7 \\
& AKS  & 76.2 & 55.1 & 44.7 & 71.7 & 61.9 \\
& Q-Frame  & 76.8 & 55.3 & 45.0 & 71.9 & 62.3 \\
\rowcolor{blue!7.5}
& CITE  & \textbf{77.2} & \textbf{55.4} & \textbf{45.4} & \textbf{72.0} & \textbf{62.5} \\
\midrule

\multirow{6}{*}{SAMA-4B}
& Baseline  & 74.0 & 52.7 & 41.5 & 71.8 & 60.0 \\
& Uniform  & 75.2 & 53.9 & 42.7 & 71.6 & 60.9 \\
& Global-CLIP  & 75.0 & 54.5 & 43.0 & 71.2 & 60.9 \\
& AKS      & 75.7 & 54.8 & 43.4 & 71.8 & 61.4 \\
& Q-Frame      & 75.9 & 55.4 & 43.8 & 71.7 & 61.7 \\
\rowcolor{blue!7.5}
& CITE         & \textbf{76.4} & \textbf{56.0} & \textbf{44.1} & \textbf{71.9} & \textbf{62.1} \\
\midrule

\multirow{6}{*}{VideoGLaMM}
& Baseline (Uniform) & 65.2 & 47.7 & 38.5 & 58.9 & 52.6 \\
& Uniform      & 66.4 & 48.6 & 39.3 & 58.8 & 53.3 \\
& Global-CLIP  & 66.9 & 49.0 & 39.8 & 58.6 & 53.6 \\
& AKS          & 67.2 & 49.1 & 40.1 & 59.0 & 53.9 \\
& Q-Frame      & 67.5 & 49.4 & 40.3 & 59.1 & 54.1 \\
\rowcolor{blue!7.5}
& CITE         & \textbf{68.6} & \textbf{49.6} & \textbf{41.3} & \textbf{59.5} & \textbf{54.8} \\
\bottomrule
\end{tabular}
}
\vspace{-2em}
\end{table}
% =======================================================================

\subsection{Ablation Study}
\label{sec:ablation}
\vspace{-0.5em}
\noindent We ablate CITE on Sa2VA-4B with (K=5), examining objective components, query-factor groups, candidate-expansion radius, and frame budget. Avg. denotes the mean over the displayed datasets.
% 
% =======================================================================
\begin{table}[!t]
\centering
\caption{
{\small
CITE component-wise analysis.
}
}
\vspace{-1em}
\label{tab:ablation_components}
\resizebox{\linewidth}{!}{
\begin{tabular}{l c c c c c c}
\toprule
Variant 
& Ref-DAVIS17 
& MeViS \texttt{val$^\texttt{u}$}
& MeViS \texttt{val}
& ReVOS
& Ref-SAV
& Avg. \\
\midrule
Uniform sampling              & 76.4 & 54.8 & 44.8 & 59.5 & 65.7 & 60.2 \\
Global query similarity       & 75.9 & 55.0 & 44.4 & 59.4 & 65.6 & 60.1 \\
\midrule
CITE w/o factor decomposition & 76.1 & 54.7 & 44.5 & 59.2 & 65.6 & 60.0 \\
CITE w/o motion evidence      & 76.7 & 55.0 & 44.9 & 59.7 & 66.0 & 60.5 \\
CITE w/o candidate expansion  & 76.6 & 55.1 & 45.0 & 59.6 & 66.1 & 60.5 \\
CITE w/o coverage gain        & 76.8 & 55.2 & 45.1 & 59.7 & 66.2 & 60.6 \\
CITE w/o uniqueness           & 77.0 & 55.2 & 45.2 & 59.8 & 66.3 & 60.7 \\
CITE w/o transition           & 76.9 & 55.1 & 45.1 & 59.8 & 66.3 & 60.7 \\
CITE w/o redundancy penalty   & 76.7 & 55.0 & 44.9 & 59.6 & 66.0 & 60.4 \\
\midrule
\rowcolor{blue!7.5}
Full CITE                     & \textbf{77.2} & \textbf{55.4} & \textbf{45.4} & \textbf{60.1} & \textbf{66.7} & \textbf{61.0} \\
\bottomrule
\end{tabular}
}
\vspace{-1em}
\end{table}
% =======================================================================
% 
\noindent \textbf{Component.}
Tab.~\ref{tab:ablation_components} shows that Full CITE attains the highest Avg. (61.0). Removing factor decomposition yields the largest drop (1.0) point), while ablating motion evidence, candidate expansion, coverage, uniqueness, transition, or redundancy reduces Avg. by (0.3)--(0.6) points, confirming their complementary contributions under (K=5).
% 
% =======================================================================
% \begin{table}[!h]
% \centering
% \caption{
% Ablation of factor groups in CITE using Sa2VA-4B.
% }
% \label{tab:ablation_factors}
% \resizebox{\linewidth}{!}{
% \begin{tabular}{l c c c c c c}
% \toprule
% Variant 
% & Ref-DAVIS17 
% & MeViS \texttt{val$^\texttt{u}$}
% & MeViS \texttt{val}
% & ReVOS
% & Ref-SAV
% & Avg. \\
% \midrule
% \rowcolor{blue!7.5}
% Full CITE                      & \textbf{77.2} & \textbf{55.4} & \textbf{45.4} & \textbf{60.1} & \textbf{66.7} & \textbf{61.0} \\
% \midrule
% w/o noun factors               & 75.8 & 53.9 & 44.0 & 58.5 & 65.2 & 59.5 \\
% w/o color/texture/material     & 76.5 & 54.6 & 44.7 & 59.2 & 66.1 & 60.2 \\
% w/o action factors             & 76.8 & 54.9 & 44.9 & 59.5 & 66.2 & 60.5 \\
% w/o direction factors          & 76.9 & 55.0 & 45.0 & 59.6 & 66.3 & 60.6 \\
% w/o action + direction factors & 76.5 & 54.6 & 44.7 & 59.1 & 66.0 & 60.2 \\
% w/o spatial relation factors   & 76.7 & 54.8 & 44.9 & 59.2 & 66.2 & 60.4 \\
% \bottomrule
% \end{tabular}
% }
% \end{table}
% =======================================================================
% 
\noindent \textbf{Factor and evidence.}
Tab.~\ref{tab:ablation_factors} shows that object-category factors are the most important single group, while appearance, action, direction, and spatial-relation factors all contribute to the final performance. This supports the factorized design of CITE.
% 
% =======================================================================
% \begin{table}[!h]
% \centering
% \caption{
% Ablation of the greedy selection objective using Sa2VA-4B.
% }
% \label{tab:ablation_objective}
% \resizebox{\linewidth}{!}{
% \begin{tabular}{l c c c c c c}
% \toprule
% Selector 
% & Ref-DAVIS17 
% & MeViS \texttt{val$^\texttt{u}$}
% & MeViS \texttt{val}
% & ReVOS
% & Ref-SAV
% & Avg. \\
% \midrule
% Top relevance                  & 75.8 & 54.7 & 44.5 & 59.2 & 65.8 & 60.0 \\
% Top uniqueness                 & 75.2 & 54.2 & 44.0 & 58.7 & 65.2 & 59.5 \\
% Coverage only                  & 76.1 & 54.9 & 44.7 & 59.4 & 65.9 & 60.2 \\
% Utility only                   & 76.6 & 55.0 & 44.9 & 59.6 & 66.1 & 60.4 \\
% Utility + coverage             & 76.9 & 55.2 & 45.2 & 59.8 & 66.4 & 60.7 \\
% \rowcolor{blue!7.5}
% Full CITE                      & \textbf{77.2} & \textbf{55.4} & \textbf{45.4} & \textbf{60.1} & \textbf{66.7} & \textbf{61.0} \\
% \bottomrule
% \end{tabular}
% }
% \end{table}
% =======================================================================
% 
% =======================================================================
\begin{table}[!t]
\centering
\begin{minipage}{0.48\linewidth}
\centering
\caption{
{\small
Factor groups in CITE.
}
}
\vspace{-2em}
\label{tab:ablation_factors}
\resizebox{\linewidth}{!}{%
\begin{tabular}{l c c c c c c}
\toprule
Variant 
& DAVIS17 
& MeViS-u
& MeViS
& ReVOS
& Ref-SAV
& Avg. \\
\midrule
\rowcolor{blue!7.5}
Full CITE          & \textbf{77.2} & \textbf{55.4} & \textbf{45.4} & \textbf{60.1} & \textbf{66.7} & \textbf{61.0} \\
\midrule
w/o noun           & 75.8 & 53.9 & 44.0 & 58.5 & 65.2 & 59.5 \\
w/o color/tex/mat  & 76.5 & 54.6 & 44.7 & 59.2 & 66.1 & 60.2 \\
w/o action         & 76.8 & 54.9 & 44.9 & 59.5 & 66.2 & 60.5 \\
w/o direction      & 76.9 & 55.0 & 45.0 & 59.6 & 66.3 & 60.6 \\
w/o action + dir   & 76.5 & 54.6 & 44.7 & 59.1 & 66.0 & 60.2 \\
w/o spatial rel    & 76.7 & 54.8 & 44.9 & 59.2 & 66.2 & 60.4 \\
\bottomrule
\end{tabular}%
}
\end{minipage}
\hfill
\begin{minipage}{0.48\linewidth}
\centering
\caption{
{\small
Selection objective.
}
}
\vspace{-1em}
\label{tab:ablation_objective}
\resizebox{\linewidth}{!}{%
\begin{tabular}{l c c c c c c}
\toprule
Selector 
& DAVIS17 
& MeViS-u
& MeViS
& ReVOS
& Ref-SAV
& Avg. \\
\midrule
Top relevance      & 75.8 & 54.7 & 44.5 & 59.2 & 65.8 & 60.0 \\
Top uniqueness     & 75.2 & 54.2 & 44.0 & 58.7 & 65.2 & 59.5 \\
Coverage only      & 76.1 & 54.9 & 44.7 & 59.4 & 65.9 & 60.2 \\
Utility only       & 76.6 & 55.0 & 44.9 & 59.6 & 66.1 & 60.4 \\
Utility + coverage & 76.9 & 55.2 & 45.2 & 59.8 & 66.4 & 60.7 \\
\rowcolor{blue!7.5}
Full CITE          & \textbf{77.2} & \textbf{55.4} & \textbf{45.4} & \textbf{60.1} & \textbf{66.7} & \textbf{61.0} \\
\bottomrule
\end{tabular}
}
\end{minipage}
\vspace{-1.5em}
\end{table}
% =======================================================================
% 
\noindent \textbf{Selection objective.}
Tab.~\ref{tab:ablation_objective} shows that ranking frames by a single criterion is suboptimal. Combining standalone utility, factor coverage, and redundancy reduction gives the best average result, supporting the design of the complementary-frames selection objective.
% 
% % =======================================================================
% \begin{table}[!h]
% \centering
% \caption{
% Sensitivity to candidate expansion radius \(r\) using Sa2VA-4B and \(K=5\).
% }
% \label{tab:radius_sensitivity}
% \begin{tabular}{l c c c c}
% \toprule
% Radius \(r\) 
% & Ref-DAVIS17 
% & MeViS \texttt{val$^\texttt{u}$}
% & MeViS \texttt{val}
% & Avg. \\
% \midrule
% \(r=0\) & 76.6 & 55.1 & 45.0 & 58.9 \\
% \rowcolor{blue!7.5}
% \(r=1\) & \textbf{77.2} & \textbf{55.4} & \textbf{45.4} & \textbf{59.3} \\
% \(r=2\) & 77.0 & 55.2 & 45.2 & 59.1 \\
% \(r=3\) & 76.6 & 54.9 & 44.9 & 58.8 \\
% \bottomrule
% \end{tabular}
% \end{table}
% % =======================================================================
% 
% % =======================================================================
% \begin{table}[!h]
% \centering
% \caption{
% Sensitivity to frame budget \(K\) using Sa2VA-4B.
% }
% \label{tab:budget_sensitivity}
% \begin{tabular}{l c c c c}
% \toprule
% Frame budget \(K\)
% & Ref-DAVIS17 
% & MeViS \texttt{val$^\texttt{u}$}
% & MeViS \texttt{val}
% & Avg. \\
% \midrule
% \(K=3\) & 75.2 & 53.8 & 44.0 & 57.7 \\
% \rowcolor{blue!7.5}
% \(K=5\) & 77.2 & 55.4 & 45.4 & 59.3 \\
% \(K=7\) & 77.5 & 55.6 & 45.6 & 59.6 \\
% \(K=9\) & \textbf{77.7} & \textbf{55.8} & \textbf{45.8} & \textbf{59.8} \\
% \bottomrule
% \end{tabular}
% \end{table}
% % =======================================================================
% 
%=========================================================================
\begin{table}[!t]
\centering
\begin{minipage}{0.48\linewidth}
\centering
\caption{Sensitivity to candidate expansion radius \(r\).}
\vspace{-1em}
\label{tab:radius_sensitivity}
\resizebox{\linewidth}{!}{%
\begin{tabular}{l c c c c}
\toprule
\(r\) 
& Ref-DAVIS17 
& MeViS \texttt{val$^\texttt{u}$}
& MeViS \texttt{val}
& Avg. \\
\midrule
\(0\) & 76.6 & 55.1 & 45.0 & 58.9 \\
\rowcolor{blue!7.5}
\(1\) & \textbf{77.2} & \textbf{55.4} & \textbf{45.4} & \textbf{59.3} \\
\(2\) & 77.0 & 55.2 & 45.2 & 59.1 \\
\(3\) & 76.6 & 54.9 & 44.9 & 58.8 \\
\bottomrule
\end{tabular}%
}
\end{minipage}
\hfill
\begin{minipage}{0.48\linewidth}
\centering
\caption{Sensitivity to frame budget \(K\).}
\vspace{-1em}
\label{tab:budget_sensitivity}
\resizebox{\linewidth}{!}{%
\begin{tabular}{l c c c c}
\toprule
\(K\)
& Ref-DAVIS17 
& MeViS \texttt{val$^\texttt{u}$}
& MeViS \texttt{val}
& Avg. \\
\midrule
\(3\) & 75.2 & 53.8 & 44.0 & 57.7 \\
\rowcolor{blue!7.5}
\(5\) & 77.2 & 55.4 & 45.4 & 59.3 \\
\(7\) & 77.5 & 55.6 & 45.6 & 59.6 \\
\(9\) & \textbf{77.7} & \textbf{55.8} & \textbf{45.8} & \textbf{59.8} \\
\bottomrule
\end{tabular}%
}
\end{minipage}
\vspace{-1em}
\end{table}
%===========================================================================
% 
% ===============================================================================================
\begin{figure}[!t]
\centering
% \vspace{-1em}
\includegraphics[width=\linewidth]{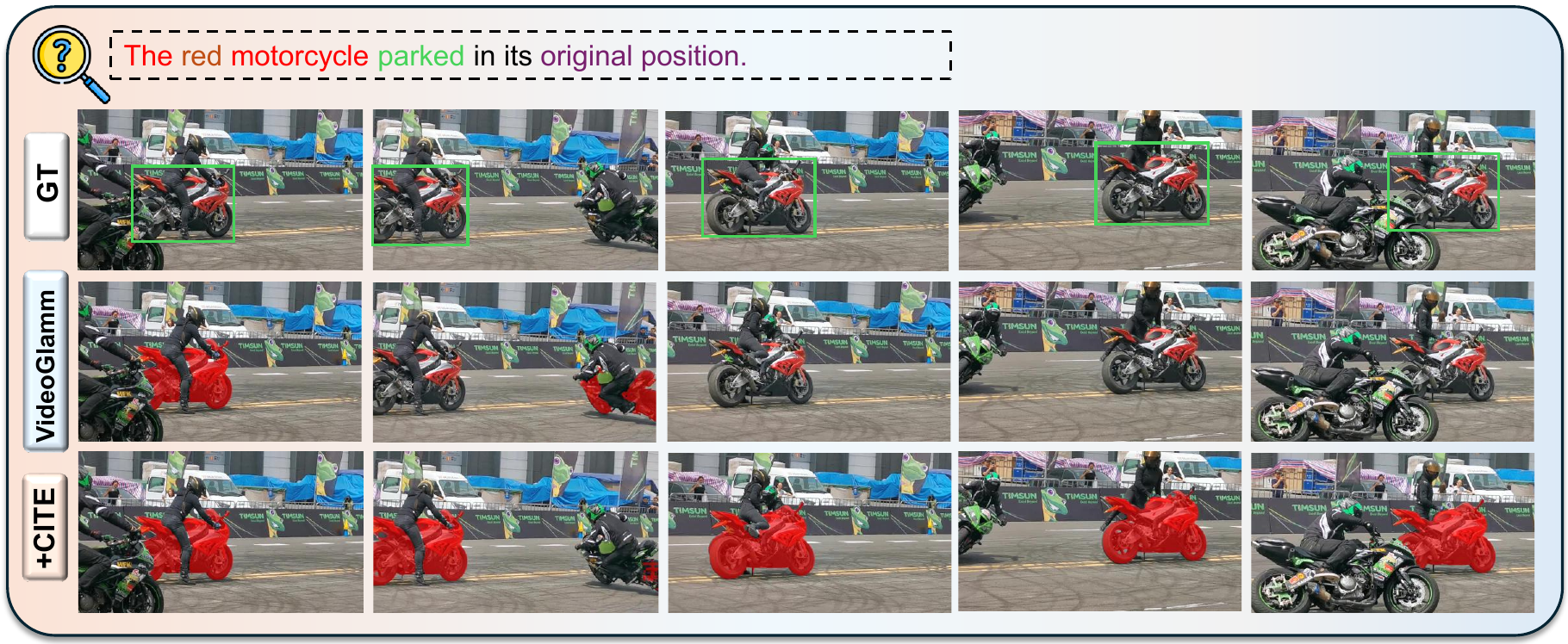}
\vspace{-2em}
\caption{
{\small
\textbf{VideoGLaMM on MeViS \texttt{val}.}
For ``\emph{The red motorcycle parked in its original position},'' CITE replaces 16-frame uniform sampling with \(K=5\) selected frames, reducing confusion with nearby motorcycles.
}
}
\label{fig:qualvidglamm}
\vspace{-0.3em}
\end{figure}
% ===============================================================================================
% 
% ===============================================================================================
\begin{figure}[!t]
\centering
\vspace{-0.7em}
\includegraphics[width=\linewidth]{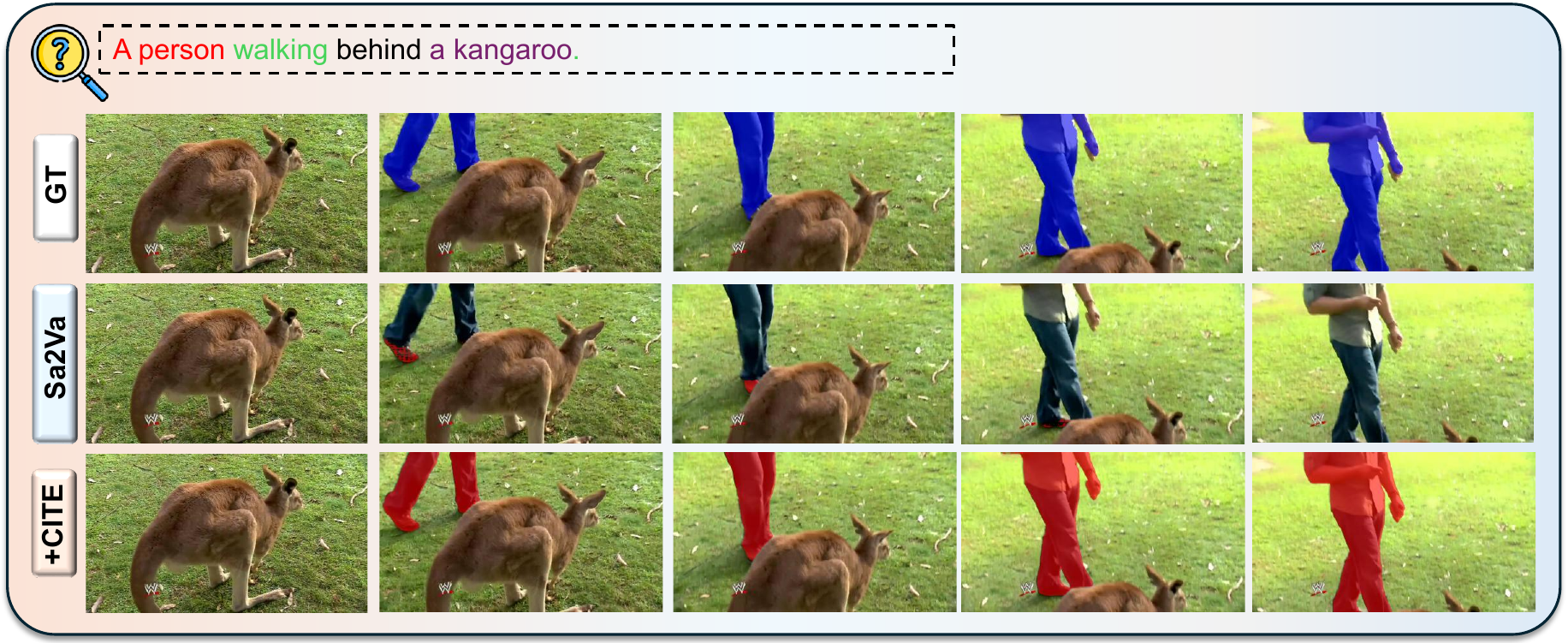}
\vspace{-2em}
\caption{
{\small
\textbf{Sa2VA on Ref-YouTube-VOS \texttt{val}.}
For ``\emph{A person walking behind a kangaroo},'' CITE selects \(K=5\) evidence frames instead of Sa2VA's first five frames, preserving later person--kangaroo relation evidence.
}
}
\label{fig:qualsa2va}
\vspace{-1em}
\end{figure}
% ===============================================================================================
% 
% ===============================================================================================
\begin{figure}[!t]
\centering
\includegraphics[width=\linewidth]{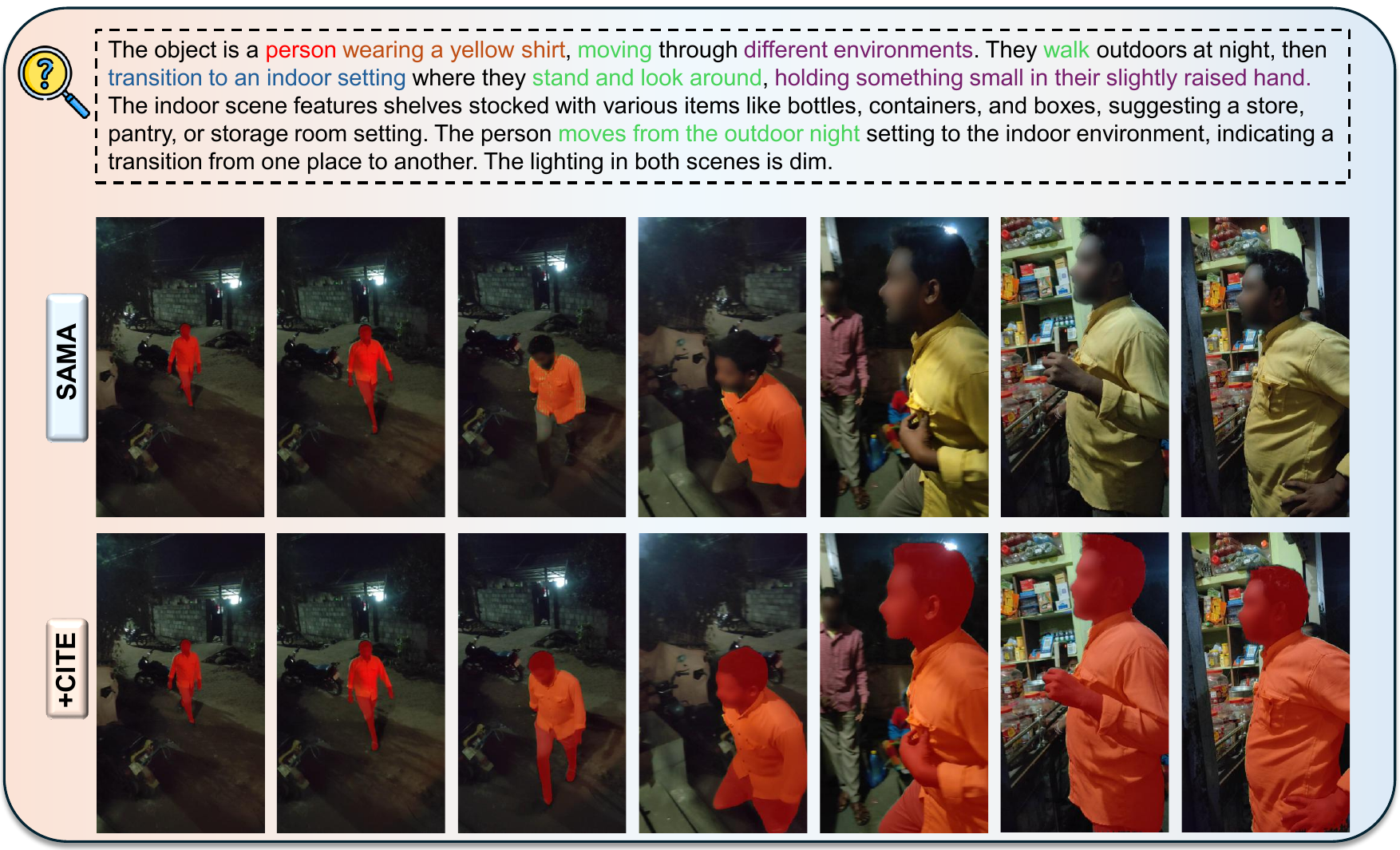}
\vspace{-2em}
\caption{
{\small
\textbf{SAMA on Ref-SAV.}
At the same \(K=5\) budget, CITE selects evidence frames instead of SAMA's first five frames, improving identity consistency and mask coverage across the outdoor-to-indoor transition.
}
}
\label{fig:qualsama}
\vspace{-1.5em}
\end{figure}
% ===============================================================================================
% ===============================================================================================
\begin{figure}[!t]
\centering
\includegraphics[width=\linewidth]{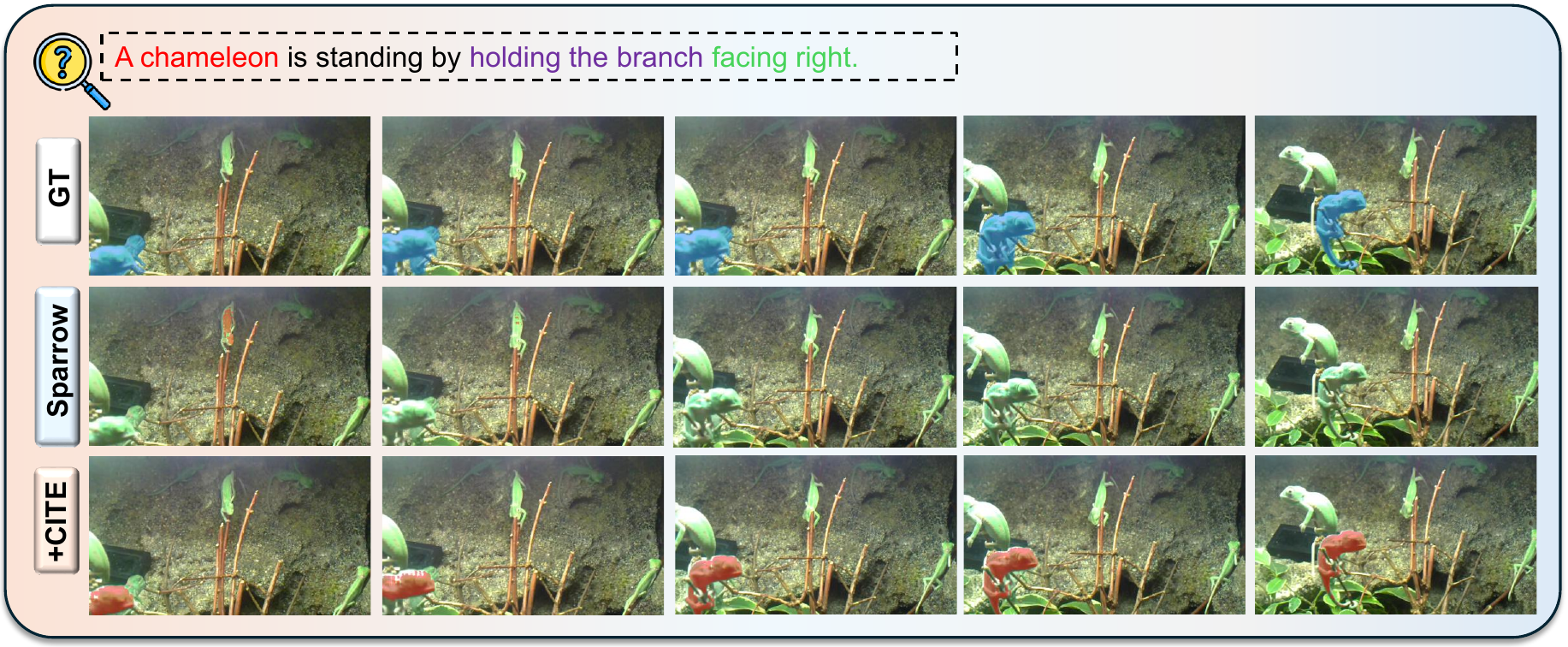}
\vspace{-2em}
\caption{
{\small
\textbf{SPARROW on Ref-YouTube-VOS \texttt{val}.}
For ``\emph{A chameleon is standing by holding the branch facing right},'' CITE replaces 16-frame uniform sampling with \(K=5\) selected frames, better preserving pose and branch-relation cues.
}
}
\label{fig:qualsparrow}
\vspace{-1.6em}
\end{figure}
% ===============================================================================================
% 
\noindent \textbf{Hyperparameter sensitivity.}
Tab.~\ref{tab:radius_sensitivity} shows that a small temporal neighborhood improves over using only factor peaks. Increasing the radius too much slightly degrades performance, suggesting that local expansion helps correct noisy peak localization, while overly broad neighborhoods introduce lower-evidence candidates.
Tab.~\ref{tab:budget_sensitivity} shows that performance saturates around \(K=9\). At \(K=5\), most of the gain is recovered with small performance increase until \(K=9\), despite requiring 80\% more reasoning frames. End-to-end latency is reported in Supp. Sec S9.

\vspace{-0.7em}
\subsection{Qualitative Analysis}
\vspace{-0.5em}
\noindent Fig.~\ref{fig:qualvidglamm}--\ref{fig:qualsparrow} show representative qualitative examples across different backbones and datasets. The default samplers often provide many frames but still miss the evidence that is decisive for the referring expression, such as the target's distinctive appearance, a transient spatial relation, a pose cue, or a scene transition. In contrast, CITE selects only five frames but biases them toward complementary query evidence. This helps the unchanged downstream backbone produce more temporally coherent masks and reduces distractor confusion.
CITE is most useful when the correct referent is determined by sparse or short-lived cues rather than by generic temporal coverage. For VideoGLaMM and SPARROW, CITE reduces confusion with visually similar distractors; for Sa2VA, it preserves later relational evidence missed by early-frame sampling; and for SAMA, it improves identity consistency across a challenging transition.
\vspace{-1.5em}
\section{Conclusion}
\label{sec:conclusion}
\noindent We introduced CITE, a training-free, plug-and-play selector that formulates keyframe selection for RVOS as query-conditioned evidence preservation. Using frozen vision-language features and lightweight temporal cues, CITE selects five complementary frames without modifying downstream pixel-grounded video MLLMs. Across multiple backbones and benchmarks, five CITE-selected frames often match or improve native-input results, with the largest gains when discriminative attributes, actions, motion, or spatial relations are sparse. Its main limitations are approximate evidence scoring, rule-based query factorization, and global motion cues. Future work will explore object-aware temporal evidence, stronger query decomposition, and adaptive frame budgets.

% \vspace{0.3em}
{\small
\section*{Acknowledgments} This research was funded by Khalifa
University of Science and Technology through the Faculty Startup Grant under
Project ID KU-INT-FSU-2026-8471000024, and supported by the Center for
Autonomous Robotic Systems, Khalifa University of Science and Technology
(KU-CARS).
}

{
    \small
    \bibliographystyle{ieeenat_fullname}
    \bibliography{egbib}
}

\clearpage
\appendix

\setcounter{section}{0}
\setcounter{subsection}{0}
\setcounter{table}{0}
\setcounter{figure}{0}
\setcounter{equation}{0}

\renewcommand{\thesection}{S\arabic{section}}
\renewcommand{\thesubsection}{S\arabic{section}.\arabic{subsection}}
\renewcommand{\thetable}{S\arabic{table}}
\renewcommand{\thefigure}{S\arabic{figure}}
\renewcommand{\theequation}{S\arabic{equation}}

% Keep hyperlink anchors distinct from those in the main paper after resetting
% the counters above.
\makeatletter
\@ifundefined{theHsection}{}{\renewcommand{\theHsection}{supp.\arabic{section}}}
\@ifundefined{theHsubsection}{}{\renewcommand{\theHsubsection}{supp.\arabic{section}.\arabic{subsection}}}
\@ifundefined{theHtable}{}{\renewcommand{\theHtable}{supp.\arabic{table}}}
\@ifundefined{theHfigure}{}{\renewcommand{\theHfigure}{supp.\arabic{figure}}}
\@ifundefined{theHequation}{}{\renewcommand{\theHequation}{supp.\arabic{equation}}}
\makeatother

\providecolor{citelight}{RGB}{232,243,252}
\providecolor{citegray}{RGB}{242,242,242}

\providecommand{\CITE}{\ifmmode\mathrm{CITE}\else\textsc{CITE}\fi}
\providecommand{\JF}{\mathcal{J}\&\mathcal{F}}
\providecommand{\tiou}{\mathrm{tIoU}}
\providecommand{\viou}{\mathrm{vIoU}}
\providecommand{\best}[1]{\textbf{#1}}
\providecommand{\datasetfive}{Ref-DAVIS17, MeViS \texttt{val}$^{\texttt{u}}$, MeViS \texttt{val}, ReVOS, and Ref-SAV}

\setlist{nosep,leftmargin=*}
\setlength{\tabcolsep}{5pt}
\renewcommand{\arraystretch}{1.12}
\emergencystretch=2em

% Start a named partial table of contents here.  Unlike \tableofcontents,
% \printcontents below reads only entries written after this point, so the
% main-paper sections (Introduction, Related Work, etc.) are excluded.
% The main WACV file already loads the required titletoc package.
\startcontents[supplement]

\begin{center}
    {\Large\bfseries Supplementary Material\par}
    \vspace{0.75em}
    {\normalsize\bfseries Contents\par}
\end{center}
\vspace{0.35em}

\begingroup
\small
\setlength{\parskip}{0pt}
% A wider label box keeps two-digit labels such as S10 separate from the title.
\titlecontents{suppsection}
    [0pt]
    {\addvspace{0.3em}\bfseries}
    {\contentslabel{2.8em}}
    {}
    {\titlerule*[0.5pc]{.}\contentspage}
% Show supplementary sections only.  Change [1] to [2] to include
% subsections as well.
\printcontents[supplement]{supp}{1}[1]{}
\endgroup
\par\addvspace{1.25em}

\subsection*{Supplementary Overview}
This supplement provides implementation details and extended analyses for
\CITE{}, a training-free keyframe selector for referring video object
segmentation (RVOS). The material separates selector-only comparisons from
default-protocol compression, evaluates alignment with fixed-budget downstream
value, reports paired same-budget gains, studies larger frame budgets and long videos,
audits the deterministic query factorizer, analyzes motion and hyperparameter
robustness, and reports end-to-end latency. Throughout, \CITE{} changes only the
reasoning frames supplied to a frozen downstream backbone. The original
referring expression, learned parameters, prompt format, mask decoder, and
tracking or propagation pipeline remain unchanged.

\section{Evaluation Conventions and Reproducibility}
\label{sec:supp_conventions}

The experiments use three deliberately distinct evaluation regimes.

\begin{enumerate}
    \item \textbf{Controlled same-budget selection.} Every selector receives
    the same frame budget $K$. This isolates the selection rule.
    \item \textbf{Default-protocol/compression.} The baseline follows its
    reproduced default inference protocol, while the $+\CITE{}$ variant uses
    $K=5$. For a
    native 16 or 36 frame backbone, this jointly measures frame selection and
    aggressive input compression.
    \item \textbf{Native-budget same-budget selection.} Default sampling,
    Q-Frame, and \CITE{} all use the same native $K\in\{16,36\}$. This isolates
    the selector at a denser operating point.
\end{enumerate}

\noindent Sa2VA and SAMA use the first five frames by default. VideoGLaMM, UniPixel, and
SPARROW use 16 uniformly sampled frames, while VideoLISA uses 36 frames. All
primary comparisons use $K=5$ for \CITE{} while explicitly labeled budget sweeps and
native-budget comparisons vary $K$. Unless explicitly stated otherwise, the
controlled sparse setting uses $K=5$.
All selected indices are sorted temporally before downstream inference, and
full-video masks are produced by the backbone's unchanged native pipeline.

% Different tables average over different benchmark sets and state the set in
% their captions: the \emph{standard four-dataset average} uses Ref-DAVIS17,
% MeViS \texttt{val}$^{\texttt{u}}$, MeViS \texttt{val}, and Ref-YouTube-VOS;
% the \emph{five-dataset ablation average} uses \datasetfive; and the
% \emph{three-dataset budget average} uses Ref-DAVIS17 and both MeViS splits.
% This distinction explains why the same Sa2VA-4B, $K=5$ configuration has
% averages of 62.5, 61.0, and 59.3 in the respective benchmark groups.

\subsection{Fixed Configuration}

Frame features are produced once per video by frozen CLIP ViT-L/14 and cached
across all expressions associated with that video. Semantic, motion, direction,
and action evidence curves are normalized temporally within the same video and
factor. Unless an ablation explicitly varies one component, the default
configuration in Table~\ref{tab:supp_configuration} is used across datasets,
backbones, and model scales.

\begin{table*}[t]
\centering
\caption{Default \CITE{} configuration. Unless explicitly varied in an
ablation, these values are shared across experiments. The
coefficients form an empirical, globally fixed evidence-preservation prior;
they are not fitted separately to any dataset or backbone.}
\label{tab:supp_configuration}
\small
\begin{tabular}{@{}p{0.23\textwidth}p{0.73\textwidth}@{}}
\toprule
Component & Configuration \\
\midrule
Frozen vision-language encoder & CLIP ViT-L/14 \\
Utility weights & $(\alpha_{\rm rel},\alpha_{\rm uniq},\alpha_{\rm gap},\alpha_{\rm trans},\alpha_{\rm ind})=(0.20,0.20,0.30,0.15,0.15)$ \\
Set objective & $\lambda_{\rm cov}=0.25$, $\lambda_{\rm red}=0.40$, and $w_g=1/|\mathcal G|$ \\
Candidate construction & Radius $r=1$, local uniqueness window $w=5$, and the top $L=\max(K,2)$ scoring frames per factor \\
Action evidence & $\operatorname{Norm}(0.7e^{\rm sem}+0.3m)$ \\
Direction gate & $\operatorname{Norm}(\max(\sigma_d,0)(0.25+m))$ \\
Transition utility & $0.4m+0.4n+0.2\Delta u_{\rm rel}$ \\
Redundancy mixture & $0.6$ temporal proximity $+\;0.4$ visual similarity \\
Fallback & Uniform sampling when no factor is extracted \\
Downstream input & Original full expression and temporally sorted selected frames \\
\bottomrule
\end{tabular}
\end{table*}

\subsection{Deterministic Query Factorization}

The factorizer is an evidence router rather than a replacement for the
downstream model's language understanding. It applies the following ordered
procedure: (i) normalize the expression while preserving common multi-word
visual phrases; (ii) detect color, texture, and material attributes from fixed
vocabularies; (iii) detect actions and motion phrases, including common verb and
\texttt{-ing} forms; (iv) detect explicit directions; (v) detect coarse spatial
relations; and (vi) assign remaining informative noun-like tokens or noun
chunks to the object group after removing stop words and already assigned
modifiers. The groups are object, appearance, action, direction, and spatial
relation; here, \emph{appearance} collectively denotes color, texture, and
material factors. The original expression is never shortened or rewritten for
RVOS inference.

For each factor, \CITE{} forms an evidence curve over the video, retains the top
$L$ scoring frames, combines the resulting indices by set union, and expands
each retained index by radius $r$. The greedy loop stops after
$\min(K,|\mathcal C|)$ selections;
the general formulation therefore permits $K'\leq\min(K,T)$. If no factor is
extracted, the selector uses the uniform fallback stated in
Table~\ref{tab:supp_configuration}.

\section{Fixed-Budget Oracle Inclusion Diagnostic}
\label{sec:supp_oracle}

\CITE{} is an evidence-preservation proxy, not a causal estimator: it does not
define a structural causal model, perform interventions, or estimate an
identifiable causal effect. We instead use a post-hoc, candidate-restricted
oracle diagnostic to test whether selected frames align with frames that have
high downstream value under a fixed budget.

For a candidate frame $t\in\mathcal C$ in a candidate set with
$|\mathcal C|>K$, its fixed-budget oracle inclusion importance is
\begin{equation}
\begin{aligned}
I_\theta(t)={}&
\max_{\substack{S\subseteq\mathcal C,\ |S|=K,\ t\in S}}
Q_\theta(S;\mathcal V,q)\\[-0.2em]
&-
\max_{\substack{S\subseteq\mathcal C,\ |S|=K,\ t\notin S}}
Q_\theta(S;\mathcal V,q),
\end{aligned}
\label{eq:supp_oracle}
\end{equation}
where $Q_\theta$ is the full-video $\JF$ score of the frozen RVOS backbone.
Ground-truth masks are used only to compute this diagnostic and never by \CITE{}.

The analysis uses Sa2VA-4B at $K=5$ on 80 video-expression pairs sampled from
Ref-DAVIS17, MeViS \texttt{val}$^{\texttt{u}}$, and Ref-YouTube-VOS. The
candidate set is the union of frames selected by Uniform, Global-CLIP, Q-Frame,
and \CITE{}, augmented with the highest-scoring unused semantic and motion
candidates. It is
capped at 17 frames after duplicate removal. The mean candidate-set size is
14.3. Across the observed candidate-size distribution, exact subset enumeration
requires approximately 184k frozen-backbone evaluations in total, or about
2.3k per expression.

Oracle Recall@5 is the within-dataset mean fraction of a selector's five frames
that belong to the five candidates with highest $I_\theta(t)$. The reported
metrics first average expressions within each dataset and then weight the three
datasets equally; they are therefore dataset-balanced rather than pooled over
all selected frames.

\begin{table*}[t]
\centering
\caption{Candidate-restricted fixed-budget oracle inclusion diagnostic using
Sa2VA-4B at $K=5$. Recall@5 is reported in percent. Lower mean rank is better.}
\label{tab:supp_oracle}
\small
\begin{tabular}{lccc}
\toprule
Selector & Oracle Recall@5 $\uparrow$ & Mean oracle rank $\downarrow$ & Mean importance $\uparrow$ \\
\midrule
Uniform & 43.2 & 5.92 & 0.17 \\
Global-CLIP & 48.9 & 5.28 & 0.23 \\
Q-Frame & 55.6 & 4.61 & 0.31 \\
\rowcolor{citelight}
\CITE{} & \best{64.7} & \best{3.82} & \best{0.44} \\
\bottomrule
\end{tabular}
\end{table*}

The initial \CITE{} marginal score has a mean Spearman correlation of 0.46 with
$I_\theta(t)$, compared with 0.32 for global full-query CLIP relevance. The
diagnostic therefore supports a deliberately limited claim: \CITE{} tends to
select frames that align better with high oracle-inclusion frames within the
candidate pool. It does not establish causal or counterfactual identification.

\section{Controlled Selector Comparisons}
\label{sec:supp_controlled}

\subsection{Comparison with Generic Relevance-Diversity Selection}

To distinguish factor-level evidence preservation from generic diversity, we
evaluate a maximal marginal relevance (MMR) selector:
\begin{equation}
\begin{aligned}
t^*=\arg\max_{t\notin S}\big[&\lambda\bar R(t)\\[-0.2em]
&-(1-\lambda)\max_{s\in S}
\bar{\operatorname{sim}}(t,s)\big],\quad \lambda=0.5.
\end{aligned}
\label{eq:supp_mmr}
\end{equation}
Relevance and visual similarity are min-max normalized per video; the first
frame is the maximum-relevance frame. Global-query MMR uses full-expression
CLIP relevance. Factor-relevance MMR uses $R(t)=M^{-1}\sum_m e_m(t)$, but omits
factor coverage, local uniqueness, transition sensitivity, factor
indispensability, candidate expansion, and the complete set objective.

\begin{table*}[t]
\centering
\caption{MMR-style comparison using Sa2VA-4B and $K=5$. The average covers
\datasetfive.}
\label{tab:supp_mmr}
\footnotesize
\resizebox{\textwidth}{!}{%
\begin{tabular}{lcccccc}
\toprule
Selector & Ref-DAVIS17 & MeViS-$u$ & MeViS & ReVOS & Ref-SAV & Avg. \\
\midrule
Global query similarity & 75.9 & 55.0 & 44.4 & 59.4 & 65.6 & 60.1 \\
Global query similarity + MMR & 76.2 & 55.1 & 44.7 & 59.6 & 66.0 & 60.3 \\
Top factor relevance & 75.8 & 54.7 & 44.5 & 59.2 & 65.8 & 60.0 \\
Factor relevance + MMR & 76.4 & 55.1 & 44.9 & 59.7 & 66.1 & 60.4 \\
Utility + coverage & 76.9 & 55.2 & 45.2 & 59.8 & 66.4 & 60.7 \\
\rowcolor{citelight}
Full \CITE{} & \best{77.2} & \best{55.4} & \best{45.4} & \best{60.1} & \best{66.7} & \best{61.0} \\
\bottomrule
\end{tabular}}
\end{table*}

Generic diversity gives a small improvement over relevance-only ranking, but
does not reproduce the full objective. This isolates the benefit of monitoring
which factor groups are covered and whether evidence is locally irreplaceable,
rather than only selecting relevant and mutually dissimilar frames.

\subsection{Paired Same-Budget Gains against Q-Frame}

For every backbone-dataset condition, we compute the paired per-expression
difference $\Delta_i=\JF_i^{\CITE}-\JF_i^{\text{Q-Frame}}$ and aggregate the
four standard datasets with equal weight. Table~\ref{tab:supp_paired_gains}
reports the resulting mean gains and the number of dataset-level wins. This is
a descriptive paired analysis; no hypothesis-test claim is made.

\begin{table*}[t]
\centering
\caption{Paired gain summary for \CITE{} versus Q-Frame under the controlled
$K=5$ setting. Mean differences use unrounded per-expression $\JF$; condition
wins count the four standard datasets.}
\label{tab:supp_paired_gains}
\small
\begin{tabular}{lcc}
\toprule
Backbone & Mean gain & Condition wins \\
\midrule
Sa2VA-1B & $+0.50$ & 4/4 \\
Sa2VA-4B & $+0.25$ & 4/4 \\
SAMA-4B & $+0.38$ & 4/4 \\
VideoGLaMM & $+0.68$ & 4/4 \\
\midrule
\rowcolor{citelight}
All controlled conditions & \best{$+0.46$} & \best{16/16} \\
\bottomrule
\end{tabular}
\end{table*}

\CITE{} wins all 16 backbone-dataset conditions. We therefore characterize the
isolated same-budget advantage as a consistent but modest empirical gain,
separately from the larger changes observed when replacing a reproduced default
protocol.

\section{Frame-Budget Analysis}
\label{sec:supp_budgets}

\subsection{Extended Budget Sweep}

\begin{table*}[t]
\centering
\caption{Accuracy-budget trade-off for \CITE{} with Sa2VA-4B. The average uses
Ref-DAVIS17 and the two MeViS splits.}
\label{tab:supp_budget}
\small
\begin{tabular}{c|cccccc}
\toprule
$K$ & 3 & 5 & 7 & 9 & 16 & 32 \\
\midrule
Ref-DAVIS17 & 75.2 & 77.2 & 77.5 & 77.7 & 77.8 & \best{77.9} \\
MeViS-$u$ & 53.8 & 55.4 & 55.6 & 55.8 & 55.9 & \best{56.0} \\
MeViS & 44.0 & 45.4 & 45.6 & 45.8 & 45.9 & \best{46.0} \\
\midrule
Average & 57.7 & 59.3 & 59.6 & 59.8 & 59.9 & \best{60.0} \\
\bottomrule
\end{tabular}
\end{table*}

Most of the small-budget gain is obtained when moving from $K=3$ to $K=5$.
Increasing from $K=5$ to $K=9$ adds about 0.4 average $\JF$ while using 80\% more
reasoning frames. Thus, $K=5$ is a low-budget knee and stress test, not a
universal optimum. A deployment can use $K=7$ or $K=9$ when the additional
memory and latency are acceptable.

\subsection{Same-Budget Selection at Native Dense Budgets}

\begin{table*}[htbp]
\centering
\caption{Native-budget same-budget selector comparison. Every method within a
backbone block uses the same $K$. Values are $\JF$; the average spans the four
standard datasets.}
\label{tab:supp_native_budget}
\footnotesize
\resizebox{\textwidth}{!}{%
\begin{tabular}{lclccccc}
\toprule
Backbone & $K$ & Sampler & Ref-DAVIS17 & MeViS-$u$ & MeViS & Ref-YT-VOS & Avg. \\
\midrule
\multirow{3}{*}{VideoGLaMM} & \multirow{3}{*}{16}
 & Default & 65.2 & 47.7 & 38.5 & 58.9 & 52.6 \\
 & & Q-Frame & 67.6 & 49.2 & 40.5 & 59.3 & 54.2 \\
 & & \CITE{} & \best{68.7} & \best{49.8} & \best{41.5} & \best{59.8} & \best{55.0} \\
\midrule
\multirow{3}{*}{SPARROW} & \multirow{3}{*}{16}
 & Default & 75.0 & 46.1 & 37.3 & 48.7 & 51.8 \\
 & & Q-Frame & 75.9 & 46.7 & 39.0 & 49.3 & 52.7 \\
 & & \CITE{} & \best{76.8} & \best{47.2} & \best{40.4} & \best{50.0} & \best{53.6} \\
\midrule
\multirow{3}{*}{VideoLISA} & \multirow{3}{*}{36}
 & Default & 65.0 & 50.1 & 39.4 & 61.2 & 53.9 \\
 & & Q-Frame & 67.6 & 50.5 & 40.0 & 61.5 & 54.9 \\
 & & \CITE{} & \best{68.7} & \best{50.8} & \best{40.5} & \best{61.8} & \best{55.5} \\
\bottomrule
\end{tabular}}
\end{table*}

At the same native budgets, \CITE{} improves over Q-Frame by 0.8, 0.9, and 0.6
average $\JF$ for VideoGLaMM, SPARROW, and VideoLISA, respectively; gains over
the default samplers are 2.4, 1.8, and about 1.5. These gains are computed from
the per-dataset cells before the displayed averages are rounded. The results
isolate selection quality from five-frame compression.

\paragraph{Architectural scope.}
\CITE{} targets backbones with an external sparse-frame reasoning or sampling
interface followed by an unchanged full-video mask pipeline. A dense
end-to-end RVOS architecture in which every clip frame participates in target
construction is not directly protocol-compatible: applying a sparse selector
would require redesigning its input schedule or adding propagation, so the
result would no longer isolate a plug-and-play selector. This scope boundary is
architectural rather than a claim that sparse selection replaces all dense
RVOS systems.

\section{Long-Video Evaluation}
\label{sec:supp_longrvos}

We evaluate on the Long-RVOS validation split. Long-RVOS contains minute-scale
sequences, occlusion, disappearance and reappearance, shot changes, and static,
dynamic, and hybrid referring expressions. In addition to spatial $\JF$, it
reports temporal IoU ($\tiou$) and spatiotemporal volume IoU ($\viou$). For a
given backbone, all selectors choose from the same decoded candidate-frame grid;
only the selection rule changes within a same-budget comparison.

\subsection{Default Protocol versus Five-Frame Compression}

\begin{table*}[htbp]
\centering
\caption{Long-RVOS default-protocol/compression results. Baselines use their
reproduced default frame budgets; every $+\CITE{}$ entry uses $K=5$. For 16- and 36-frame
backbones, this is an aggressive compression test rather than a selector-only
comparison.}
\label{tab:supp_long_sparse}
\scriptsize
\resizebox{\textwidth}{!}{%
\begin{tabular}{l c ccc ccc ccc}
\toprule
\multirow{2}{*}{Backbone} & \multirow{2}{*}{Native $K$}
 & \multicolumn{3}{c}{Default} & \multicolumn{3}{c}{$+\CITE{}$, $K=5$} & \multicolumn{3}{c}{Change} \\
\cmidrule(lr){3-5}\cmidrule(lr){6-8}\cmidrule(lr){9-11}
 & & $\JF$ & $\tiou$ & $\viou$ & $\JF$ & $\tiou$ & $\viou$ & $\Delta\JF$ & $\Delta\tiou$ & $\Delta\viou$ \\
\midrule
Sa2VA-1B & 5 & 38.4 & 68.8 & 31.9 & 40.2 & 69.3 & 33.6 & +1.8 & +0.5 & +1.7 \\
Sa2VA-4B & 5 & 41.7 & 69.6 & 35.2 & 43.8 & 70.1 & 37.3 & +2.1 & +0.5 & +2.1 \\
Sa2VA-8B & 5 & 43.4 & 70.0 & 36.9 & 45.0 & 70.5 & 38.5 & +1.6 & +0.5 & +1.6 \\
Sa2VA-26B & 5 & 45.8 & 70.7 & 39.2 & 47.9 & 71.3 & 41.4 & +2.1 & +0.6 & +2.2 \\
\midrule
\rowcolor{citegray}
VideoLISA & 36 & 14.0 & 69.0 & 8.1 & 13.6 & 68.8 & 7.8 & -0.4 & -0.2 & -0.3 \\
\midrule
SAMA-1B & 5 & 39.2 & 68.7 & 32.5 & 41.4 & 69.3 & 34.7 & +2.2 & +0.6 & +2.2 \\
SAMA-4B & 5 & 43.0 & 69.8 & 36.5 & 44.7 & 70.3 & 38.1 & +1.7 & +0.5 & +1.6 \\
SAMA-8B & 5 & 44.6 & 70.3 & 38.0 & 46.0 & 70.7 & 39.3 & +1.4 & +0.4 & +1.3 \\
\midrule
UniPixel-3B & 16 & 41.9 & 70.2 & 35.5 & 43.4 & 70.7 & 37.0 & +1.5 & +0.5 & +1.5 \\
UniPixel-7B & 16 & 42.8 & 70.6 & 36.4 & 44.0 & 71.0 & 37.6 & +1.2 & +0.4 & +1.2 \\
VideoGLaMM & 16 & 31.5 & 68.7 & 26.8 & 34.0 & 69.4 & 29.5 & +2.5 & +0.7 & +2.7 \\
SPARROW & 16 & 28.4 & 68.9 & 24.1 & 29.7 & 69.3 & 25.5 & +1.3 & +0.4 & +1.4 \\
\midrule
\rowcolor{citelight}
Macro-average & -- & 37.9 & 69.6 & 31.8 & \best{39.5} & \best{70.1} & \best{33.4} & \best{+1.6} & \best{+0.5} & \best{+1.6} \\
\bottomrule
\end{tabular}}
\end{table*}

The macro-average improves by 1.6 $\JF$, 0.5 $\tiou$, and 1.6 $\viou$. The
VideoLISA row provides an important counterexample: compressing its native 36
frames to five reduces all three metrics slightly. This indicates that a model
trained or designed around dense temporal context can lose useful continuity
under an overly aggressive budget.

\subsection{Native-Budget Same-Budget Long-RVOS Results}

\begin{table*}[htbp]
\centering
\caption{Long-RVOS native-budget same-budget comparison. Triplets are
$\JF/\tiou/\viou$. Every method in a block uses the same $K$.}
\label{tab:supp_long_native}
\small
\begin{tabular}{lc l ccc}
\toprule
Backbone & $K$ & Sampler & $\JF$ & $\tiou$ & $\viou$ \\
\midrule
\multirow{3}{*}{VideoLISA} & \multirow{3}{*}{36}
 & Default & 14.0 & 69.0 & 8.1 \\
 & & Q-Frame & 14.7 & 69.3 & 8.9 \\
 & & \CITE{} & \best{15.8} & \best{69.7} & \best{10.2} \\
\midrule
\multirow{3}{*}{UniPixel-3B} & \multirow{3}{*}{16}
 & Default & 41.9 & 70.2 & 35.5 \\
 & & Q-Frame & 42.8 & 70.5 & 36.5 \\
 & & \CITE{} & \best{44.1} & \best{71.0} & \best{37.9} \\
\midrule
\multirow{3}{*}{UniPixel-7B} & \multirow{3}{*}{16}
 & Default & 42.8 & 70.6 & 36.4 \\
 & & Q-Frame & 43.5 & 70.8 & 37.2 \\
 & & \CITE{} & \best{44.6} & \best{71.2} & \best{38.4} \\
\midrule
\multirow{3}{*}{VideoGLaMM} & \multirow{3}{*}{16}
 & Default & 31.5 & 68.7 & 26.8 \\
 & & Q-Frame & 33.0 & 69.1 & 28.4 \\
 & & \CITE{} & \best{34.8} & \best{69.6} & \best{30.3} \\
\midrule
\multirow{3}{*}{SPARROW} & \multirow{3}{*}{16}
 & Default & 28.4 & 68.9 & 24.1 \\
 & & Q-Frame & 29.4 & 69.2 & 25.1 \\
 & & \CITE{} & \best{30.5} & \best{69.6} & \best{26.4} \\
\bottomrule
\end{tabular}
\end{table*}

At VideoLISA's native $K=36$, \CITE{} improves over the default by 1.8 $\JF$,
0.7 $\tiou$, and 2.1 $\viou$. The native-budget recovery supports the
interpretation that its five-frame degradation is at least partly a compression
effect. At $K=16$, \CITE{} improves $\JF$ over the default by 2.2, 1.8, 3.3,
and 2.1 for UniPixel-3B, UniPixel-7B, VideoGLaMM, and SPARROW, respectively.

\section{Deterministic Factorizer Analysis}
\label{sec:supp_parser}

\subsection{Coverage and Manual Audit}

Full-dataset coverage measures whether at least one factor is extracted; it is
not a correctness metric. The manual audit samples 50 expressions from each of
six benchmark strata. Two annotators independently label an overlapping set of
120 expressions, reaching Cohen's $\kappa=0.86$ on factor-group labels. A parse
is usable when it contains the main target-object cue and introduces no severe
contradictory factor.

\begin{table*}[htbp]
\centering
\caption{Deterministic factorizer analysis. Panel (a) is a manual audit;
precision, recall, F1, and group accuracy are percentages. Panel (b) is
full-dataset nonempty-extraction coverage.}
\label{tab:supp_parser_audit}
\small
\centering
\textbf{(a) Manual audit on 300 randomly sampled expressions.}\par\smallskip
\resizebox{\textwidth}{!}{%
\begin{tabular}{lrrrrrr}
\toprule
Dataset/split & \#Expr. & Usable parse & Precision & Recall & F1 & Group acc. \\
\midrule
Ref-DAVIS17 & 50 & 96.0 & 94.1 & 88.7 & 91.3 & 95.4 \\
MeViS \texttt{val}$^{\texttt{u}}$ & 50 & 94.0 & 92.0 & 86.4 & 89.1 & 93.7 \\
MeViS \texttt{val} & 50 & 94.0 & 92.6 & 86.9 & 89.7 & 94.0 \\
Ref-YouTube-VOS & 50 & 96.0 & 94.5 & 89.2 & 91.8 & 95.6 \\
ReVOS & 50 & 92.0 & 90.1 & 84.0 & 86.9 & 91.8 \\
Ref-SAV & 50 & 94.0 & 91.3 & 85.6 & 88.4 & 92.7 \\
\midrule
\rowcolor{citelight}
Overall & 300 & \best{94.3} & \best{92.4} & \best{86.8} & \best{89.5} & \best{93.9} \\
\bottomrule
\end{tabular}}

\vspace{0.8em}
\centering
\textbf{(b) Full-dataset coverage and factor statistics.}\par\smallskip
\begin{tabular}{lrrrr}
\toprule
Dataset & Parsed (\%) & Fallback (\%) & Avg. \# factors & Multi-group (\%) \\
\midrule
Ref-DAVIS17 & 98.2 & 1.8 & 2.48 & 73.5 \\
MeViS \texttt{val}$^{\texttt{u}}$ & 97.6 & 2.4 & 2.91 & 80.4 \\
MeViS \texttt{val} & 97.8 & 2.2 & 2.84 & 79.1 \\
Ref-YouTube-VOS & 98.0 & 2.0 & 2.32 & 68.7 \\
ReVOS & 96.5 & 3.5 & 3.16 & 84.9 \\
Ref-SAV & 97.1 & 2.9 & 2.68 & 76.5 \\
\midrule
\rowcolor{citelight}
Macro-average & \best{97.5} & 2.5 & 2.73 & 77.2 \\
\bottomrule
\end{tabular}
\end{table*}

The main audit errors are missed unusual actions or nested relations rather
than severe false-positive factors. Since the downstream model still receives
the original full expression, a missed factor reduces selection benefit but
does not delete language information from segmentation.

\subsection{Robustness to Linguistic Perturbations}

\begin{table*}[t]
\centering
\caption{Robustness to deterministic factor extraction with Sa2VA-4B at
$K=5$. Factor-dropout results average three random seeds. The average spans
\datasetfive.}
\label{tab:supp_factor_robustness}
\footnotesize
\resizebox{\textwidth}{!}{%
\begin{tabular}{lcccccc}
\toprule
Variant & Ref-DAVIS17 & MeViS-$u$ & MeViS & ReVOS & Ref-SAV & Avg. \\
\midrule
Global query similarity & 75.9 & 55.0 & 44.4 & 59.4 & 65.6 & 60.1 \\
\CITE{} without factor decomposition & 76.1 & 54.7 & 44.5 & 59.2 & 65.6 & 60.0 \\
Alternative lexical templates & 77.1 & 55.3 & 45.3 & 60.0 & 66.6 & 60.9 \\
20\% factor dropout & 77.0 & 55.2 & 45.2 & 59.9 & 66.5 & 60.8 \\
40\% factor dropout & 76.7 & 55.0 & 44.9 & 59.7 & 66.2 & 60.5 \\
\rowcolor{citelight}
Full \CITE{} & \best{77.2} & \best{55.4} & \best{45.4} & \best{60.1} & \best{66.7} & \best{61.0} \\
\bottomrule
\end{tabular}}
\end{table*}

Alternative templates reduce the average by only 0.1 point, while 20\% and
40\% factor dropout reduce it by 0.2 and 0.5. Factorization is therefore useful
without requiring a perfect parse on the evaluated benchmarks.

\subsection{Expressions with Semantic and Motion Cues}

The semantic-motion overlap subset contains at least one object, appearance, or
spatial cue and at least one action or direction cue. It covers 44.1\%, 62.4\%,
60.8\%, 66.2\%, and 55.6\% of Ref-DAVIS17, MeViS
\texttt{val}$^{\texttt{u}}$, MeViS \texttt{val}, ReVOS, and Ref-SAV,
respectively (57.8\% on average).

\begin{table*}[htbp]
\centering
\caption{Performance on expressions containing both semantic and
motion/action cues using Sa2VA-4B and $K=5$. Values are $\JF$.}
\label{tab:supp_multicue}
\footnotesize
\resizebox{\textwidth}{!}{%
\begin{tabular}{lcccccc}
\toprule
Method & Ref-DAVIS17 & MeViS-$u$ & MeViS & ReVOS & Ref-SAV & Avg. \\
\midrule
Uniform & 75.0 & 53.9 & 43.9 & 58.4 & 64.8 & 59.2 \\
Global query similarity & 74.6 & 54.1 & 43.7 & 58.2 & 64.6 & 59.0 \\
\CITE{} without factor decomposition & 74.9 & 53.8 & 43.8 & 58.2 & 64.7 & 59.1 \\
\CITE{} without motion & 75.6 & 54.5 & 44.6 & 58.9 & 65.4 & 59.8 \\
Alternative lexical templates & 76.2 & 54.9 & 44.9 & 59.4 & 65.8 & 60.2 \\
20\% factor dropout & 76.1 & 54.8 & 44.8 & 59.3 & 65.7 & 60.1 \\
40\% factor dropout & 75.8 & 54.6 & 44.6 & 59.0 & 65.2 & 59.8 \\
\rowcolor{citelight}
Full \CITE{} & \best{76.4} & \best{55.0} & \best{45.0} & \best{59.6} & \best{66.0} & \best{60.4} \\
\bottomrule
\end{tabular}}
\end{table*}

On multi-cue expressions, full \CITE{} improves over Uniform by 1.2 average
$\JF$, over global similarity by 1.4, and over the no-factor variant by 1.3.
Removing motion reduces the average by 0.6, showing that temporal evidence is
useful precisely when both cue types are present.

\subsection{Illustrative Factorization Failures}

\begin{table*}[t]
\centering
\caption{Illustrative limitations of the deterministic factorizer.}
\label{tab:supp_parser_failures}
\small
\begin{tabular}{@{}p{0.18\textwidth}p{0.34\textwidth}p{0.42\textwidth}@{}}
\toprule
Failure type & Example expression & Likely effect \\
\midrule
Nested relation & ``the person left of the man behind the car'' & Relations may be extracted without their hierarchy. \\
Negation & ``the dog not wearing a leash'' & ``Leash'' may be treated as positive evidence. \\
Ordinal/counting & ``the second person from the left'' & The current groups do not model ordinality explicitly. \\
Rare synonym or metaphor & ``the vehicle darting across the road'' & An uncommon action phrase may be missed. \\
Coreference & ``the man who picks it up'' & The referent of ``it'' may remain unresolved. \\
Nonvisual cue & ``the person who seems confused'' & The cue may require higher-level interpretation. \\
\bottomrule
\end{tabular}
\end{table*}

Dependency or constituency parsers and open-vocabulary span extractors are
compatible drop-in alternatives because the selection objective needs only
factor texts and coarse groups. Such extensions can improve irregular-prompt
coverage but are outside the deterministic configuration evaluated here.

\section{Motion and Hyperparameter Robustness}
\label{sec:supp_robustness}

\subsection{Motion-Cue Diagnostics}

The motion signal is a weak temporal prior rather than optical flow, tracking,
or object-conditioned motion. The high camera/background-motion subset is the
top quartile within each dataset under a combined score based on homography
corner displacement and residual frame difference after compensation.

\begin{table*}[t]
\centering
\caption{Targeted motion robustness using Sa2VA-4B at $K=5$. All entries are
average $\JF$.}
\label{tab:supp_motion_targeted}
\small
\begin{tabular}{lccc}
\toprule
Setting & Uniform & \CITE{} without motion & Full \CITE{} \\
\midrule
Five-dataset ablation average & 60.2 & 60.5 & \best{61.0} \\
Action/direction queries & 59.6 & 60.1 & \best{60.8} \\
High camera/background motion & 57.7 & 58.4 & \best{58.9} \\
\bottomrule
\end{tabular}
\end{table*}

Full \CITE{} remains strongest on the high-motion subset, but only by 0.5 point
over the no-motion variant. This supports the intended interpretation that
motion is complementary rather than dominant; it does not imply robustness to
all forms of ego-motion or background dynamics.

\begin{table}[t]
\centering
\caption{Alternative lightweight motion designs. The average spans
\datasetfive.}
\label{tab:supp_motion_alternatives}
\small
\begin{tabular}{lc}
\toprule
Motion design & Avg. $\JF$ \\
\midrule
No motion evidence & 60.5 \\
Motion magnitude only & 60.6 \\
Direction without magnitude weighting & 60.6 \\
Local blockwise frame difference & 60.8 \\
\rowcolor{citelight}
Default \CITE{} & \best{61.0} \\
\bottomrule
\end{tabular}
\end{table}

\subsection{Objective-Weight Sensitivity}

The semantic-biased and dynamic-biased variants assign a $2{:}1$ relative
factor-group weight to semantic versus dynamic groups, or the reverse, and then
renormalize over the groups present. The utility-coefficient variants use the
order $(\alpha_{\rm rel},\alpha_{\rm uniq},\alpha_{\rm gap},
\alpha_{\rm trans},\alpha_{\rm ind})$: equal coefficients use
$(0.20,0.20,0.20,0.20,0.20)$, and dynamic-cue-biased utility uses
$(0.15,0.15,0.20,0.30,0.20)$.

% A relevance-biased utility row is intentionally excluded: the available
% source tables disagree on its ReVOS and Ref-SAV cells, and no raw log was
% supplied to resolve the conflict.

\begin{table*}[htbp]
\centering
\caption{Sensitivity to factor-group and top-level objective weights with
Sa2VA-4B at $K=5$. Only the listed choice changes; the average spans
\datasetfive.}
\label{tab:supp_weights}
\footnotesize
\resizebox{\textwidth}{!}{%
\begin{tabular}{lcccccc}
\toprule
Setting & Ref-DAVIS17 & MeViS-$u$ & MeViS & ReVOS & Ref-SAV & Avg. \\
\midrule
Semantic-biased factor weights & 77.3 & 55.0 & 45.1 & 59.8 & 66.4 & 60.7 \\
Dynamic-biased factor weights & 76.8 & 55.6 & 45.6 & 60.0 & 66.2 & 60.8 \\
Equal utility coefficients & 77.0 & 55.2 & 45.2 & 59.9 & 66.5 & 60.8 \\
Dynamic-cue-biased utility & 76.9 & 55.5 & 45.5 & 60.0 & 66.3 & 60.8 \\
$\lambda_{\rm cov}=0.10$ & 76.9 & 55.1 & 45.1 & 59.7 & 66.3 & 60.6 \\
$\lambda_{\rm cov}=0.50$ & 77.0 & 55.3 & 45.3 & 60.0 & 66.5 & 60.8 \\
$\lambda_{\rm red}=0.20$ & 76.8 & 55.0 & 44.9 & 59.6 & 66.1 & 60.5 \\
$\lambda_{\rm red}=0.60$ & 77.0 & 55.2 & 45.3 & 59.9 & 66.4 & 60.8 \\
\rowcolor{citelight}
Default \CITE{} & 77.2 & 55.4 & 45.4 & \best{60.1} & \best{66.7} & \best{61.0} \\
\bottomrule
\end{tabular}}
\end{table*}

Every tested average lies within 0.5 point of the default. Semantic-biased
weights slightly favor Ref-DAVIS17, whereas dynamic-biased choices slightly
favor MeViS; neither improves the cross-dataset average. The coefficients are
therefore locally stable in the tested range, although no theoretical
optimality is claimed.

\subsection{Auxiliary-Constant Sensitivity}

\begin{table*}[htbp]
\centering
\caption{Sensitivity to auxiliary constants. Only the listed quantity changes;
the average spans \datasetfive.}
\label{tab:supp_auxiliary}
\footnotesize
\begin{tabular}{lll}
\toprule
Quantity & Setting & Avg. $\JF$ \\
\midrule
Candidate radius & $r=0,\ \mathbf{1},\ 2,\ 3$ & $60.5,\ \mathbf{61.0},\ 60.8,\ 60.6$ \\
Uniqueness window & $w=3,\ \mathbf{5},\ 7,\ 9$ & $60.8,\ \mathbf{61.0},\ 60.9,\ 60.7$ \\
Top-scoring frames & $L=\lceil K/2\rceil,\ \mathbf{\max(K,2)},\ 2K,\ 3K$ & $60.5,\ \mathbf{61.0},\ 60.9,\ 60.7$ \\
Action mix & $(0.9,0.1),\ \mathbf{(0.7,0.3)},\ (0.5,0.5),\ (0.3,0.7)$ & $60.7,\ \mathbf{61.0},\ 60.8,\ 60.4$ \\
Direction gate offset & $0.10,\ \mathbf{0.25},\ 0.50$ & $60.7,\ \mathbf{61.0},\ 60.8$ \\
Transition $(m,n,\Delta u)$ & $(0.6,0.2,0.2),\ \mathbf{(0.4,0.4,0.2)},\ (0.2,0.6,0.2),\ (0.3,0.3,0.4)$ & $60.7,\ \mathbf{61.0},\ 60.8,\ 60.6$ \\
Redundancy (temporal, visual) & $(0.8,0.2),\ \mathbf{(0.6,0.4)},\ (0.4,0.6)$ & $60.7,\ \mathbf{61.0},\ 60.8$ \\
\bottomrule
\end{tabular}
\end{table*}

Nearby auxiliary settings remain within 0.6 average point. Larger motion
weights eventually reduce performance, consistent with treating global motion
as an auxiliary signal rather than the dominant evidence source. These are
five-dataset averages; the candidate-radius table in the main paper averages
only Ref-DAVIS17 and the two MeViS splits, so its numerical averages differ.

\section{When Evidence-Preserving Selection Helps}
\label{sec:supp_peakiness}

To characterize the temporal concentration of query evidence, define
\begin{equation}
\bar e(t)=\frac{1}{M}\sum_{m=1}^{M}e_m(t),\qquad
\rho=\frac{\max_t\bar e(t)}{\epsilon+T^{-1}\sum_{t=1}^{T}\bar e(t)}.
\label{eq:supp_peakiness}
\end{equation}
High $\rho$ indicates sparse or temporally localized evidence; low $\rho$
indicates persistent evidence. Ref-YouTube-VOS is split into the top and bottom
thirds of $\rho$ without changing any selector parameter. Results average the
four backbones in the controlled $K=5$ comparison.

\begin{table*}[t]
\centering
\caption{Ref-YouTube-VOS performance by temporal evidence concentration. All
selectors use $K=5$; values are average $\JF$.}
\label{tab:supp_peakiness}
\small
\begin{tabular}{lcccccc}
\toprule
Subset & Fraction & Uniform & Global-CLIP & AKS & Q-Frame & \CITE{} \\
\midrule
Sparse/peaked & 33.3\% & 64.8 & 64.7 & 65.0 & 65.4 & \best{66.0} \\
Persistent/diffuse & 33.3\% & 69.7 & 69.5 & 69.8 & 69.9 & \best{70.0} \\
\midrule
All examples & 100\% & 67.4 & 67.1 & 67.6 & 67.7 & \best{67.9} \\
\bottomrule
\end{tabular}
\end{table*}

On sparse/peaked examples, \CITE{} improves over Uniform by 1.2 and Q-Frame by
0.6. On persistent/diffuse examples, the corresponding gains shrink to 0.3 and
0.1. This directly supports the restrained mechanism claim that evidence
preservation has more headroom when decisive cues are temporally localized.
Across the 12 main-table backbones, the average improvements are 2.35 on
Ref-DAVIS17, 1.88 on MeViS \texttt{val}$^{\texttt{u}}$, 2.62 on MeViS
\texttt{val}, and 0.53 on Ref-YouTube-VOS. The latter still improves in 11 of
12 backbone settings; excluding the 36-to-5 VideoLISA compression case gives a
mean gain of 0.70.

\section{End-to-End Latency and Deployment Trade-offs}
\label{sec:supp_runtime}

The runtime table in the main paper reports selector preprocessing overhead
only. This section gives a separate full-pipeline diagnostic so selector cost
is not conflated with downstream reasoning, segmentation, or propagation.

Cached selector cost for a video with $T$ frames and $Q$ expressions is
\begin{equation}
C_{\rm sel}(T,Q)=\frac{C_{\rm CLIP}^{\rm img}(T)}{Q}+C_{\rm query}.
\label{eq:supp_selector_cost}
\end{equation}
For a 100-frame video, $C_{\rm CLIP}^{\rm img}=1.21$ s and
$C_{\rm query}=0.03$ s, giving 1.24 s per expression for $Q=1$ and 0.27 s for
$Q=5$. Timing uses a single A100 GPU. Total time includes selector preprocessing,
downstream reasoning/segmentation on the input frames, and full-video
propagation or tracking.

\begin{table*}[htbp]
\centering
\caption{End-to-end latency on 100-frame videos. Times are seconds per
expression. Accuracy is the four-standard-dataset average, reported alongside
the timing rather than computed only on the timed videos.}
\label{tab:supp_runtime}
\footnotesize
\resizebox{\textwidth}{!}{%
\begin{tabular}{llrrrrrr}
\toprule
Backbone & Setting & Frames & Selector & Reason/seg. & Prop. & Total & Avg. $\JF$ \\
\midrule
\multirow{3}{*}{Sa2VA-4B}
 & Default & 5 & 0.00 & 2.84 & 3.18 & 6.02 & 59.9 \\
 & $+\CITE{}$, $Q=1$ & 5 & 1.24 & 2.84 & 3.18 & 7.26 & \best{62.5} \\
 & $+\CITE{}$, $Q=5$ & 5 & 0.27 & 2.84 & 3.18 & 6.29 & \best{62.5} \\
\midrule
\multirow{3}{*}{VideoGLaMM}
 & Default & 16 & 0.00 & 8.96 & 3.70 & 12.66 & 52.6 \\
 & $+\CITE{}$, $Q=1$ & 5 & 1.24 & 3.05 & 3.70 & 7.99 & \best{54.8} \\
 & $+\CITE{}$, $Q=5$ & 5 & 0.27 & 3.05 & 3.70 & 7.02 & \best{54.8} \\
\midrule
\multirow{3}{*}{VideoLISA}
 & Default & 36 & 0.00 & 14.76 & 4.12 & 18.88 & 53.9 \\
 & $+\CITE{}$, $Q=1$ & 5 & 1.24 & 2.91 & 4.12 & 8.27 & \best{54.2} \\
 & $+\CITE{}$, $Q=5$ & 5 & 0.27 & 2.91 & 4.12 & 7.30 & \best{54.2} \\
\bottomrule
\end{tabular}}
\end{table*}

For native-five-frame Sa2VA-4B, \CITE{} is an accuracy-latency trade-off rather
than a speed-up: accuracy rises from 59.9 to 62.5, but one-query time rises from
6.02 s to 7.26 s. For denser native protocols, the selector cost is offset by
fewer expensive reasoning frames. VideoGLaMM changes from 12.66 s/52.6 to
7.99 s/54.8, and VideoLISA from 18.88 s/53.9 to 8.27 s/54.2 (time/accuracy).

Let $B$ be the native input budget and $c_{\rm RVOS}$ the mean downstream
reasoning cost per frame. Assuming reasoning cost scales linearly with the
number of input frames and propagation cost is unchanged, compressing to five
frames reduces latency when
\begin{equation}
C_{\rm sel}(T,Q)<(B-5)c_{\rm RVOS}.
\label{eq:supp_breakeven}
\end{equation}

\begin{table*}[t]
\centering
\caption{Per-frame downstream cost above which \CITE{} is latency-beneficial on
100-frame videos.}
\label{tab:supp_breakeven}
\small
\begin{tabular}{lcc}
\toprule
Native budget & $Q=1$ & $Q=5$ \\
\midrule
$B=16$ & $c_{\rm RVOS}>0.113$ s/frame & $c_{\rm RVOS}>0.025$ s/frame \\
$B=36$ & $c_{\rm RVOS}>0.040$ s/frame & $c_{\rm RVOS}>0.009$ s/frame \\
\bottomrule
\end{tabular}
\end{table*}

The least favorable case is a short, single-query video processed by a native
five-frame backbone. Under the same linear scaling model, a 32-frame video with
$Q=1$ is estimated to change from 3.86 s to 4.28 s for Sa2VA-4B and from
10.14 s to 4.65 s for VideoGLaMM, whose reasoning input still falls from 16
frames to five. These estimates motivate an explicit deployment rule: bypass
\CITE{} when latency is the sole priority and
the video is short, has one query, and the backbone already uses five frames;
prefer it for denser native budgets, shared cached video features, or sparse
decisive evidence.

\section{Limitations and Claim Boundaries}
\label{sec:supp_limitations}

The results support \CITE{} as a fixed, training-free evidence-preservation
selector, subject to the following boundaries.

\begin{itemize}
    \item \textbf{No causal claim.} The implemented score is a model-agnostic
    proxy; the oracle analysis is candidate-restricted and post-hoc.
    \item \textbf{Approximate motion.} Global differences can be corrupted by
    camera motion, background dynamics, or several moving objects.
    \item \textbf{Approximate language routing.} Negation, relation hierarchy,
    counting, coreference, metaphor, and nonvisual cues remain difficult.
    \item \textbf{Empirical coefficients.} The fixed constants are locally
    stable in the reported sweeps but are not theoretically optimal.
    \item \textbf{Finite temporal context.} A small $K$ can miss gradual state
    changes, event ordering, disappearance/reappearance, or long-range identity.
    Backbones requiring dense continuity may need a larger or adaptive budget.
    \item \textbf{Evidence-dependent headroom.} Gains are smaller when target
    evidence is temporally persistent and already captured by simple sampling.
    \item \textbf{Training-free trade-off.} The selector transfers without
    keyframe supervision or backbone-specific fitting, but it cannot learn a
    dataset-specific policy from downstream feedback.
    \item \textbf{Latency is conditional.} \CITE{} is not universally faster,
    particularly for short, single-query, native-five-frame deployments.
\end{itemize}

Natural extensions include object-aware temporal evidence, camera-motion
compensation, stronger drop-in factor extractors, learned or validation-adaptive
weights, and a budget policy that allocates additional frames when evidence is
diffuse or the downstream backbone requires temporal continuity.

\stopcontents[supplement]

\end{document}